\documentclass{article}

\usepackage{listings}
\usepackage[table]{xcolor}
\usepackage{tcolorbox}
\usepackage{subcaption}

\newtcolorbox{promptbox-case}[1][]{
    colback=green!3,
    colbacktitle=green!50!black,
    coltitle=white,
    title=#1,
    fonttitle=\bfseries,
    boxsep=5pt,
    left=0pt,
    right=0pt,
    top=0pt,
    bottom=0pt,
    boxrule=1pt,
    arc=3mm,
    toptitle=3pt,
    bottomtitle=3pt
}

\newtcolorbox{promptbox}[1][]{
    colback=gray!10,
    colbacktitle=black!60,
    coltitle=white,
    title=#1,
    fonttitle=\bfseries,
    boxsep=5pt,
    left=0pt,
    right=0pt,
    top=0pt,
    bottom=0pt,
    boxrule=1pt,
    arc=3mm,
    toptitle=3pt,
    bottomtitle=3pt
}

\usepackage{listings}  
\usepackage{fancyvrb}  
\usepackage{times}
\usepackage{latexsym}
\usepackage{multirow} 
\usepackage[ruled,vlined,linesnumbered]{algorithm2e}
\usepackage{booktabs}
\usepackage{multirow}
\usepackage{adjustbox}
\usepackage{array}
\usepackage{amsmath}

\usepackage[T1]{fontenc}
\usepackage[utf8]{inputenc}
\usepackage{microtype}
\usepackage{amsmath}   
\usepackage{amssymb} 
\usepackage{inconsolata}
\usepackage{graphicx}
\usepackage{hyperref}
\usepackage[accepted]{icml2025}
\usepackage{pifont}

\definecolor{darkgreen}{RGB}{0,100,0}
\definecolor{lightgreen}{rgb}{0.881, 0.936, 0.859}
\definecolor{darkred}{RGB}{139,0,0}
\definecolor{lightyellow}{rgb}{1, 0.9, 0.9}

\usepackage{amsmath}
\usepackage{amssymb}
\usepackage{mathtools}
\usepackage{amsthm}
\usepackage{inconsolata}
\newtheorem{takeaway}{Take-away}
\usepackage[capitalize,noabbrev]{cleveref}

\theoremstyle{plain}

\theoremstyle{definition}

\theoremstyle{remark}

\usepackage[textsize=tiny]{todonotes}

\icmltitlerunning{Submission and Formatting Instructions for ICML 2026}

\begin{document}

\twocolumn[
\icmltitle{R\textsuperscript{2}-Write: Reflection and Revision for Open-Ended Writing with Deep Reasoning}



\icmlsetsymbol{equal}{*}

\begin{icmlauthorlist}
\icmlauthor{Wanlong Liu}{equal,comp}
\icmlauthor{Bo Zhang}{comp}
\icmlauthor{Chenliang Li}{comp}
\icmlauthor{Shaopeng Lai}{comp}
\icmlauthor{Yuning Wu}{comp}
\icmlauthor{Xuanyu Lei}{sch}
\icmlauthor{Ming Yan}{comp}
\end{icmlauthorlist}


\icmlaffiliation{comp}{Tongyi Lab, Alibaba Group}

\icmlaffiliation{sch}{Tsinghua University, Beijing, China}

\icmlcorrespondingauthor{Ming Yan}{ym119608@alibaba-inc.com}

\icmlkeywords{Machine Learning, ICML}
\vskip 0.3in
]


\printAffiliationsAndNotice{\icmlEqualContribution} 

\begin{abstract}
While deep reasoning with long chain-of-thought has dramatically improved large language models in verifiable domains like mathematics, its effectiveness for open-ended tasks such as writing remains unexplored. In this paper, we conduct a systematic investigation revealing that existing mainstream reasoning models achieve limited gains on open-ended writing tasks. Our further analysis shows that these models lack deep reflection and revision patterns in open-ended writing, resulting in substantially smaller improvements compared to mathematical reasoning tasks.
To address this limitation,
we introduce \textbf{R\textsuperscript{2}-Write}, an automated framework that synthesizes high-quality thinking trajectories enriched with explicit reflection and revision patterns through iterative writer-judge interaction. To prevent redundant reflections, we design a process reward mechanism that supervises reflection quality during reinforcement learning, improving both performance and token efficiency. Extensive experiments across multiple creative writing and deep-research benchmarks demonstrate significant improvements, validating that explicitly incorporating reflection and revision patterns unlocks deep reasoning capabilities for open-ended writing tasks.
\end{abstract}

\section{Introduction}

The emerging paradigm of \emph{deep reasoning} is reshaping the reasoning capabilities of large language models (LLMs), marking a pivotal step in their evolution and driving substantial performance gains in verifiable domains~\cite{fu2025deep, jaech2024openai}. Grounded in the test-time scaling paradigm, deep reasoning unlocks advanced abilities such as multi-step planning, complex problem solving, and reflective self-correction~\cite{gandhi2025cognitive}, leading to remarkable improvements in areas like mathematics~\cite{liu2025qfft, guan2025rstar, yu2025z1}. These advances are largely propelled by reinforcement learning with verifiable rewards (RLVR), where clear reward signals for correct outcomes guide the model’s search through vast solution spaces~\cite{zhang2025survey, guo2025deepseek}.


However, the effectiveness of {deep reasoning} in open-ended domains remains largely unexplored. Open-ended writing, which has no definitive answers, has become an increasingly crucial capability for LLMs in real-world applications, such as professional report writing, novel composition, legal drafting, and educational content creation~\cite{yao2019plan, que2024hellobench, wu2025writingbench, wu2025shifting}. The absence of verifiable ground truth makes it difficult to directly apply the RLVR paradigm to the open-ended writing tasks. Although some recent works~\cite{jia2025writing, wu2025longwriter, lei2025writing} attempt to bridge this gap by constructing rubric-based evaluation schemes and using LLM judges to produce numerical reward signals, a fundamental question remains unanswered: \textit{What gains can deep reasoning actually bring to open-ended writing tasks?}

To address this question, we first empirically evaluate existing leading reasoning models on open-ended writing tasks. Surprisingly, we discover that {deep reasoning} provides only limited performance gains, far below the remarkable improvements observed in mathematical tasks. To explore this gap, we perform a fine-grained analysis of the reasoning trajectories and characterize the thinking patterns. Compared with mathematical reasoning, we find that open-ended writing is dominated by ``planning-oriented'' patterns, whereas ``verification'' and ``backtracking'' patterns—key drivers of success in mathematics—are severely underrepresented.

\begin{table*}[t]
\centering
\caption{Performance of different models on creative writing and mathematical reasoning benchmarks.
For Qwen3-30B-A3B, \emph{No Thinking} corresponds to \texttt{Qwen3-30B-A3B-Instruct} and \emph{Thinking} to \texttt{Qwen3-30B-A3B-Thinking}. For Qwen3-8B, \emph{Thinking} and \emph{No Thinking} use the same base model with different system prompts.
$\Delta$ denotes the relative percentage change from the left column to the right column.}
\vskip 0.05in
\resizebox{1\textwidth}{!}{
\begin{tabular}{lccccccccc}
\toprule
\multirow{2}{*}{\textbf{Benchmark}} &
\multicolumn{3}{c}{\textbf{Qwen3-30B-A3B}} &
\multicolumn{3}{c}{\textbf{Qwen3-8B}} &
\multicolumn{3}{c}{\textbf{DeepSeek}} \\
\cmidrule(lr){2-4} \cmidrule(lr){5-7} \cmidrule(lr){8-10}
& \textbf{No Thinking} & \textbf{Thinking} & $\boldsymbol{\Delta}$ (\%) 
& \textbf{No Thinking} & \textbf{Thinking} & $\boldsymbol{\Delta}$ (\%) 
& \textbf{V3.1} & \textbf{R1-0528} & $\boldsymbol{\Delta}$ (\%) \\
\midrule
WritingBench  & 78.2 & 79.1 & {\color{darkgreen}{+1.2}}
              & 71.2 & 71.8 & {\color{darkgreen}{+0.8}}
              & 78.8 & 79.0 & {\color{darkgreen}{+0.3}} \\
HelloBench    & 79.8 & 80.4 & {\color{darkgreen}{+0.7}}
              & 71.9 & 72.2 & {\color{darkgreen}{+0.4}}
              & 77.4 & 78.0 & {\color{darkgreen}{+0.7}} \\
MATH 500      & 88.2 & 97.5 & {\color{darkgreen}{+10.5}}
              & 86.8 & 96.4 & {\color{darkgreen}{+11.1}}
              & 92.2 & 98.8 & {\color{darkgreen}{+7.2}} \\
AIME 25       & 24.6 & 73.9 & {\color{darkgreen}{+200.4}}
              & 20.9 & 67.3 & {\color{darkgreen}{+221.5}}
              & 43.9 & 87.5 & {\color{darkgreen}{+99.3}} \\
\bottomrule
\end{tabular}}

\label{tab:writing_math_comparison}
\end{table*}

This observation motivates us to revisit the nature of writing tasks more carefully.  For open-ended writing tasks, the determinants of response quality fall into two categories: (1) \textit{unverifiable creative aspects}, such as esthetic judgments and cultural preferences, which lack objective notions of right or wrong; (2) \textit{verifiable aspects}\footnote{ The ``verifiable'' here differs from rule-based verification in mathematics. While open-ended writing lacks ground truth answers, these errors can still be identified through LLM judges.}, such as query misalignment, quality defects, citation mistakes, repetition, and factual inaccuracies. Crucially, we assume that verifiable errors can be identified and corrected through reflection and revision, analogous to error correction in mathematical reasoning. This motivates us to explicitly incorporate \emph{reflection} and \emph{revision} patterns into the model's reasoning process for writing tasks.

In this paper, we propose an automated data construction framework that synthesizes high-quality thinking trajectories enriched with \emph{reflection} and \emph{revision} patterns. Our framework employs a writer model and a judge model: the writer model produces an initial answer, the judge model provides iterative feedback, and the writer model internalizes the judge's comments as its own reflective thoughts and refines the answer accordingly. Using this approach, we synthesize 3K high-quality deep reasoning data for SFT and 5K data for RL. To encourage effective and precise reflection patterns, we also design a \textit{process reward mechanism} that explicitly supervises reflection and revision quality during RL, substantially reducing redundant reflections while improving performance and token efficiency. Our approach achieves significant improvements on multiple representative creative writing and deep-research benchmarks.

Our contributions can be summarized as follows:
\begin{itemize}
    \item We provide the first systematic analysis of test-time scaling for open-ended writing, revealing that current reasoning models lack effective self-reflection and revision mechanisms, resulting in substantially smaller gains compared to mathematical reasoning.

    \item We propose R\textsuperscript{2}-Write, a scalable automated pipeline that synthesizes high-quality thinking trajectories enriched with reflection and revision patterns. During RL training, we introduce a process reward mechanism that explicitly supervises reflection quality, ensuring effective and efficient pattern usage.

    \item Extensive experiments demonstrate significant improvements across multiple writing and deep-research benchmarks, with detailed analyses validating the effectiveness of explicitly incorporating reflection and revision patterns for open-ended writing tasks.
\end{itemize}

\begin{figure*}[ht!]
  \centering
  \vspace{-5pt}
  \resizebox{0.95\textwidth}{!}{
  \includegraphics[width=1.0\textwidth]{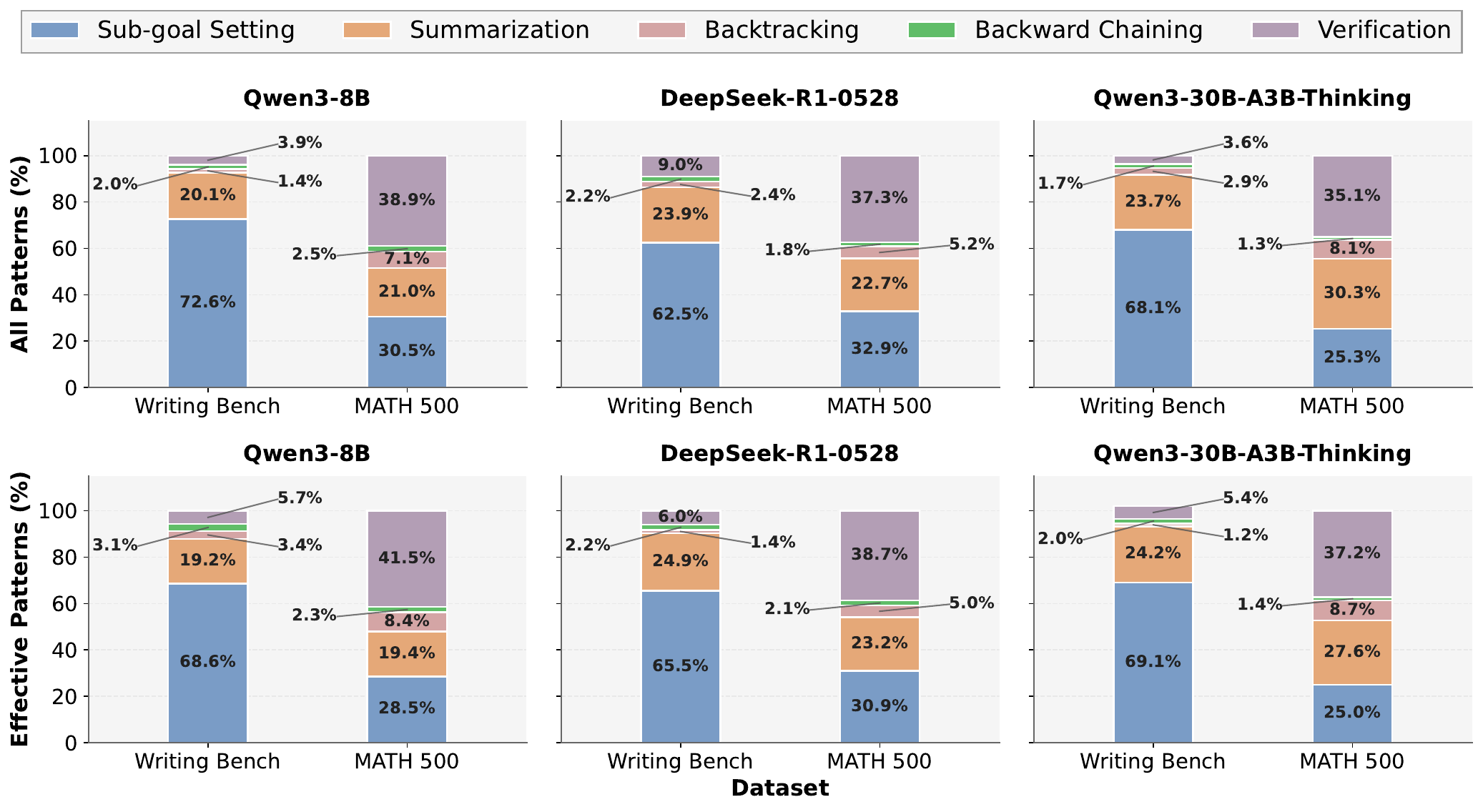}
  }
\caption{
Thinking pattern analysis.
The first row shows the pattern distributions of three reasoning models on WritingBench and MATH500.
The second row reports, for each model and task, the proportion of patterns that are judged to be helpful for obtaining the correct (or high-scoring) answer.
All pattern annotations are obtained using Claude-4.5-Sonnet~\cite{anthropic2025claude}.
\label{fig:Thinking Pattern analysis results}
}

\end{figure*}

\section{Pilot Study}
To answer the question of what gains deep reasoning can bring to open-ended writing, we first evaluate its performance on creative writing benchmarks (Section \ref{sec:2.1}), then analyze the underlying thinking patterns to understand the differences (Section \ref{sec:2.2}).

\subsection{How Does Deep Reasoning Work for Writing Tasks}
\label{sec:2.1}
Many existing studies have shown that \emph{deep reasoning} brings substantial improvements on verifiable tasks such as mathematics. However, its benefits for open-ended writing remain largely underexplored. In this work, we systematically evaluate the impact of deep reasoning on creative writing tasks.

\paragraph{Experiment Setting.}
We consider two creative writing benchmarks, {WritingBench}~\cite{wu2025writingbench} and {HelloBench}~\cite{que2024hellobench}, and evaluate representative open-source reasoning models, including the Qwen3 series~\cite{yang2025qwen3} and DeepSeek-R1~\cite{guo2025deepseek}, against their non-thinking counterparts. Detailed evaluation settings are provided in the Appendix~\ref{appendix: 2.1}.

\paragraph{Results.} As shown in Table~\ref{tab:writing_math_comparison}, deep reasoning improves model performance on WritingBench and HelloBench, but the gains are much smaller than those observed on mathematical reasoning tasks such as MATH500~\cite{lightman2023let} and AIME 25~\cite{aime2025}.

\begin{takeaway}
Deep reasoning provides gains on open-ended writing tasks, but these improvements are noticeably smaller than those observed in mathematical reasoning.
\end{takeaway}

\subsection{Thinking Patterns Analysis for Open-ended Writing}
\label{sec:2.2}
While deep reasoning yields modest gains on open-ended writing, the underlying causes of its limited effectiveness remain unclear. We conduct a more fine-grained and comprehensive analysis, aiming to uncover  regularities in the reasoning traces that may explain why deep reasoning does not have substantial improvements for creative writing.

\paragraph{Experiment Setting.}
Prior works~\cite{gandhi2025cognitive, zhang2025critique} categorize thinking patterns into five major types, as summarized in Table~\ref{tab:define_reasoning_patterns} in Appendix~\ref{appendix: 2.2}. 
Following this setup, we employ Claude-4.5-Sonnet~\cite{anthropic2025claude} to extract and classify these patterns in the reasoning traces of two representative open-source reasoning model families, including the Qwen3~\cite{yang2025qwen3} and DeepSeek~\cite{guo2025deepseek} series, on both mathematical reasoning and open-ended writing tasks. 
We further assess whether each identified pattern contributes positively to obtaining the correct (or high-quality) answer, following the method of~\cite{zhang2025critique}. 
Detailed experimental settings and prompts are provided in the Appendix~\ref{appendix: 2.2}.

\paragraph{Results.} As shown in Figure~\ref{fig:Thinking Pattern analysis results}, on MATH500, the ``Answer Verification'' and ``Backtracking'' patterns play a crucial role in helping models obtain correct (high quality) answers. 
In contrast, on WritingBench the models predominantly exhibit the ``Subgoal Setting'' pattern, while ``Verification'' and ``Backtracking'' are largely absent. 

\begin{takeaway}
Verification and backtracking patterns substantially help reasoning models obtain correct answers on math task, yet these patterns are severely lacking in open-ended writing tasks.
\end{takeaway}

\begin{figure*}[ht!]
  \centering
  \vspace{-5pt}
  \resizebox{0.97\textwidth}{!}{
  \includegraphics[width=1.0\textwidth]{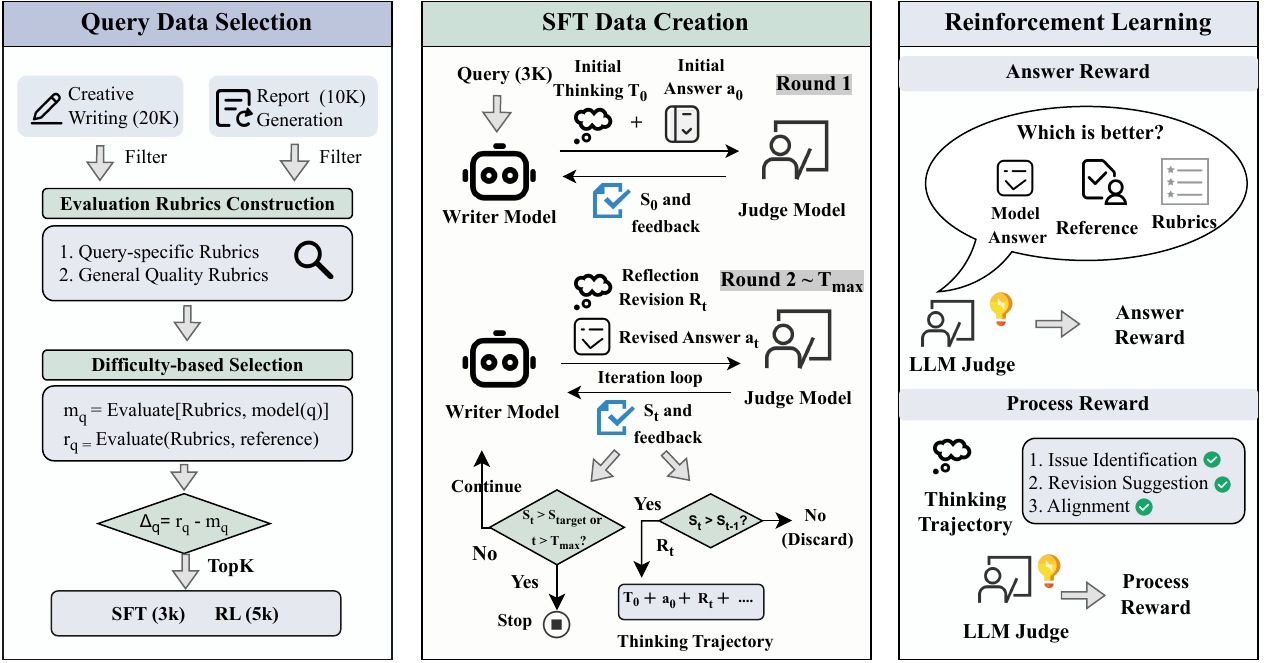}
  }
\caption{
Overview of the R\textsuperscript{2}-Write pipeline, which consists of three main parts: query data selection, data creation and RL.
\label{fig:method}
}
\end{figure*}

\section{Methodology}

In this work, we aim to enrich the reasoning trajectories  with reflection and revision patterns that are currently absent in open-ended writing tasks. We define \textbf{reflection} as \textit{the process of identifying issues in the current draft through verification or backtracking}, and \textbf{revision} as \textit{proposing and implementing concrete solutions to address these issues}. As shown in Figure~\ref{fig:method}, our methodology consists of constructing high-quality training data enriched with these patterns, followed by SFT and RL to effectively learn and apply them.




\subsection{Query Data Selection}
\label{sec:3.1}
To ensure our method generalizes effectively across different writing scenarios, we curate a diverse dataset encompassing multiple types of open-ended writing tasks.  Specifically, we include both creative-writing queries and report-generation tasks, where the latter corresponds to deep-research writing scenarios in which users provide specific requirements along with collected materials to generate structured, comprehensive reports.

For creative writing, we use the \textsc{DeepWriting} subset (20K instances) from~\cite{wang2025reverse}, a high-quality dataset covering over 14 categories.  
For report generation, we {collect 10K deep-research queries from public online websites~\cite{team2025tongyi}, spanning 10 major categories\footnote{For each query, we retrieve relevant content using a search engine and include it as supplementary context.}.} The detailed pipeline is shown in Appendix~\ref{appendix: Query Section}.

\paragraph{Evaluation Rubrics Construction.}  Since open-ended writing tasks lack ground-truth answers, we design evaluation rubrics that align with human writing standards.
Following prior evaluation work~\cite{coelho2025deepresearchgymfreetransparentreproducible, futuresearch2025deepresearchbenchevaluating}, for creative writing, we design two types of evaluation rubrics using Claude-4.5-Sonnet for every query: \emph{query-specific rubrics} and \emph{general quality rubrics}. Query-specific rubrics define fine-grained, task-dependent criteria, focusing on coverage of key information required by the query. General quality rubrics assess overall answer quality, such as fluency, completeness, and creativity. To ensure the quality of evaluation rubrics, we conducted quality verification and alignment experiments to verify their consistency with human judgment. (details in Appendix~\ref{appendix: Evaluation Rubrics Construction}.)

\paragraph{Difficulty-Based Query Selection.}
To maximize data efficiency, we identify challenging queries with substantial performance gaps, where the model most requires reflection and revision.
 For each query $q$, we generate a response using Qwen3-30B-A3B~\cite{yang2025qwen3} and compute rubric-based scores. Let $S^q_r$ and $S^q_m$ denote the aggregated scores of the reference answer and the model's answer, respectively, combining both query-specific and generic evaluation dimensions. We define the performance gap as:
\[
\Delta_q = S^q_r - S^q_m,
\]
and select the top-$k$ queries with the largest gaps. This difficulty-based filtering yields 3K high-quality instances for SFT (in Section \ref{sec:3.2}) and 5K instances for RL (in Section \ref{sec:rl}), ensuring our training focuses on genuinely challenging cases that maximize model improvement.

\subsection{SFT data Creation Pipeline}
\label{sec:3.2}
We construct a high-quality SFT dataset in which the thinking process explicitly exhibits reflection and revision patterns. Our goals are as two fold:
\vspace{-2mm}
\begin{itemize}
\vspace{-1mm}
    \item \textbf{Pattern enrichment.} Introduce new patterns while also preserving existing valuable patterns (e.g., subgoal setting), so that the model acquires a richer and more diverse repertoire of reasoning strategies rather than simply replacing one pattern with another.
    \item \textbf{Effective reflection and revision.} Ensure these patterns are triggered appropriately. For example, reflection is triggered only when there are genuine issues in the draft, and revision leads to targeted revisions that improve the quality of the final answer.
    
\end{itemize}

Based on these goals, we design an efficient end-to-end \emph{writer–judge} data synthesis pipeline. The writer model first generates an initial answer to a given query.  The judge model then evaluates this answer based on the rubrics, providing both a numerical score and detailed feedback. Finally, the writer model performs self-refinement based on the judge's feedback.

Each self-refinement iteration consists of two steps.  \textbf{Step 1:} The writer model first generates a thinking segment by internalizing the judge's feedback into its own thinking process of identifying errors. The thinking segment includes both \textit{reflection}---explicitly verifying and identifying issues in the current answer---and \textit{revision}---proposing concrete solutions to address these issues.  During this step, we deliberately inject human-like thinking patterns. The prompts explicitly encourage phrases that signify cognitive exploration and self-reflection, such as "Hmm...maybe I should revise..." or "Wait, I found that...", triggering a more natural self-reflection style  while avoiding rigid, mechanical, and formulaic thinking. \textbf{Step 2:} the writer model generates a revised answer conditioned on the reflection from Step 1. 

The judge model then re-evaluates the revised answer with a new score and feedback.
\textbf{We retain revisions whose scores improve over the previous ones, ensuring that each reflection–revision cycle is effective}. {The iterative process terminates when either the cumulative score of 
the current generated answer exceeds the predefined threshold, or the 
number of iterations reaches the maximum allowed limit.} This iterative loop ultimately produces training examples in which 
reflection and revision are explicitly articulated within the thinking 
trace and are grounded in verifiable, incremental quality improvements 
(see Appendix~\ref{sec: Data Construction Pipeline} for more details).

\subsection{Reinforcement Learning}
\label{sec:rl}

After SFT on the constructed data in Section~\ref{sec:3.2}, the model acquires basic reflection and revision capabilities. To further strengthen its ability to use these patterns effectively, we adopt a RL stage.

Following recent work~\cite{lei2025writingrladvancinglongformwriting, wu2025longwriter}, we use Proximal Policy Optimization (PPO)~\cite{schulman2017proximal} as RL algorithm and adopt an LLM-as-a-Judge paradigm to provide rewards. Specifically, we employ pairwise comparison between the model's output and a high-quality reference\footnote{Reference selection is detailed in the Appendix~\ref{appendix: Evaluation Rubrics Construction}.}, evaluated based on the rubrics constructed in Section~\ref{sec:3.1}. This pairwise setup offers more discriminative feedback for open-ended writing, better capturing subtle quality differences and directions for improvement~\cite{lei2025writing}.

However, most existing approaches attach rewards only to the final answer and ignore the thinking process. This is misaligned with our goal of teaching the model to use ``reflection'' and ``revision'' appropriately. We therefore assign rewards to both the thinking trajectory and the final answer, encouraging the model to trigger these patterns only when necessary and to use them in a correct  way.

\paragraph{Reward Design.}
We decompose the total reward into a process-level and an answer-level component:
\begin{equation}
    R_{\text{all}} = 
    \begin{cases}
        \alpha \, R_{\text{a}} + (1-\alpha) \, R_{\text{p}}, & \text{if } R_{\text{a}} > 0, \\
        R_{\text{a}}, & \text{otherwise,}
    \end{cases}
\end{equation}
where \(0 < \alpha < 1\) is a weighting coefficient hyperparameter. Here, \(R_{\text{a}}\) measures the final answer quality via a pairwise LLM-as-a-Judge model, while \(R_{\text{p}}\) evaluates how the thinking trajectory uses ``reflection'' and ``revision''. {Notably, to mitigate the risk of reward hacking, we compute the 
process reward $R_{\text{p}}$ \textbf{only when} $R_{\text{a}} > 0$, 
thereby preventing the model from being rewarded for well-structured 
reasoning processes that nonetheless yield incorrect final answers.}

\begin{table*}[t]
\centering
\caption{Performance comparison across different training methods on Writing Bench, Deepresearch Gym, and DiscoX datasets. $\uparrow$ indicates higher is better, $\downarrow$ indicates lower is better. $\text{RL}_\text{p}$ denotes RL training with process-level supervision.} 
\vskip 0.05in
\resizebox{1\textwidth}{!}{
\begin{tabular}{lcccccccc}
\toprule
\multirow{2}{*}{\textbf{Method}} & \multirow{2}{*}{\textbf{Training}} & \multicolumn{1}{c}{\textbf{Writing Bench}} & \multicolumn{3}{c}{\textbf{Deepresearch Gym}} & \multicolumn{3}{c}{\textbf{DiscoX}} \\
\cmidrule(lr){3-3} \cmidrule(lr){4-6} \cmidrule(lr){7-9}
& & Score($\uparrow$) & KPR($\uparrow$) & KPC($\downarrow$) & Quality($\uparrow$) & Acc.($\uparrow$) & Fluency($\uparrow$) & Appro.($\uparrow$) \\
\midrule

\multicolumn{9}{c}{\textbf{Proprietary LLMs}} \\
\midrule
GPT-5 & -  & 85.87 & 76.26 & 2.20 & 7.98 & 42.62 & 12.42 & 12.40 \\

Gemini2.5-Pro & -  & 82.15  & 73.20 & 2.68 & 7.40 & 40.23 & 11.86 & 13.26\\

Claude-4-sonnet & -  & 81.76    & 77.10 & 1.87 & 7.86 & 38.65 & 11.42 & 11.80 \\
\midrule

\multicolumn{9}{c}{\textbf{SFT-Based Methods}} \\
\midrule
Qwen3-4B & -  & 67.40 & 60.04 & 2.88  & 5.76  & 12.82 & 7.70 & 8.20 \\
Qwen3-8B & - & 71.84 & 63.78 & 2.62 & 5.98 & 18.80 & 8.74 & 8.80 \\
Suri-7B & SFT & 49.70 & - & - & - & - & - & - \\
Longwriter-9B & SFT & 68.20 & 57.22 & 5.98 & 5.66 & 14.12  & 7.20 & 7.88 \\
Reverse-Engineering-8B & SFT & 73.10 & 60.01 & 4.89 & 6.02 & 18.62 & 9.58 & 9.86 \\
\rowcolor{lightyellow}  Qwen3-8B + Distill & SFT & 78.82 & 66.37 & 2.48 & 6.60 & 21.70 & 9.30 & 10.00 \\
\rowcolor{lightyellow} \text{Qwen3-8B + R\textsuperscript{2}-Write-Last} & SFT & 76.70 & 65.72 & 2.62 & 6.60 & 20.14 & 9.20 & 9.24 \\
\rowcolor{lightgreen} Qwen3-4B + R\textsuperscript{2}-Write-SFT  & SFT  & 78.20 & 67.12 & 2.68 & 6.44 & 18.22 & 8.48 & 9.40 \\
\rowcolor{lightgreen} Qwen3-8B + R\textsuperscript{2}-Write-SFT  & SFT & \textbf{80.71} & \textbf{70.05} & \textbf{2.33} & \textbf{7.19} & \textbf{22.70} & \textbf{10.06} & \textbf{10.90} \\
 
\midrule
\multicolumn{9}{c}{\textbf{RL-Based Methods}} \\
\midrule
Qwen3-4B + RL & PPO  & 74.48 & 64.20 & 2.84 & 6.23 & 15.26 & 7.86 & 8.96 \\

Qwen3-8B + RL & PPO & 78.42 & 68.84 & 2.73 & 6.86 & 19.30 & 9.53 & 9.95 \\
Qwen3-8B + Distill + RL & SFT + PPO & 80.22 & 68.92 & 2.70 & 7.25 & 20.42 & 9.84 & 10.20 \\
Writing-RL-7B & SFT + PPO & 78.20 & 68.10 & 3.62 & 7.05  & 15.20 & 7.70 & 7.20 \\
LongWriter-Zero-32B & SFT + GRPO & 76.56 & 68.20 & 3.24 & 7.05 & 19.20 & 9.74 & 9.20 \\
\rowcolor{lightgreen} Qwen3-4B + R\textsuperscript{2}-Write-SFT + $\text{RL}_\text{p}$ & SFT + PPO &  80.22 & 69.72 & 2.42 & 7.10 & 18.64 & 10.22 & 10.28 \\
\rowcolor{lightgreen} Qwen3-8B + R\textsuperscript{2}-Write-SFT + RL & SFT + PPO & 82.22 & 71.63 & 2.24 & 7.46 & 23.16 & \textbf{10.91} & \textbf{10.95} \\
\rowcolor{lightgreen} Qwen3-8B + R\textsuperscript{2}-Write-SFT + $\text{RL}_\text{p}$ & SFT + PPO & \textbf{83.80} & \textbf{72.50} & \textbf{2.10} & \textbf{7.80} & \textbf{23.50} & \text{10.82} & \text{10.90} \\
\bottomrule
\end{tabular}
}

\label{tab:main results}
\end{table*}

\paragraph{Process Reward.} For each training instance, we leverage an LLM to identify and extract all reflection segments $\mathcal{M}$ from the thinking trajectory where the model explicitly recognizes a problem and proposes a solution.
\begin{equation}
    \mathcal{M} = \{ M_{F_1}, \dots, M_{F_K} \}.
\end{equation}
Each \(M_{F_i}\) is judged along three aspects:

\begin{itemize}
    \item \textbf{Issue Identification} \(R_{\text{find}}^{(i)} \in \{-1, +1\}\): valuable / spurious issue.
    \item \textbf{Revision Suggestion} \(R_{\text{rev}}^{(i)} \in \{-1, +1\}\): revision idea violates / matches the rubrics.
    \item \textbf{Execution Alignment} \(R_{\text{align}}^{(i)} \in \{-1, +1\}\): whether the final answer implements the planned revision.
\end{itemize}

For each reflection segment \(M_{F_i}\), we compute:
\begin{equation}
    R_{\text{p}}^{(i)} = 
    \begin{cases}
        +1, & \text{if } R_{\text{find}}^{(i)} > 0 \text{ and } R_{\text{rev}}^{(i)} > 0 \text{ and } R_{\text{align}}^{(i)} > 0, \\
        -1, & \text{otherwise.}
    \end{cases}
\end{equation}
The overall process reward is the average over all reflection segments:
\begin{equation}
    R_{\text{p}} = \frac{1}{K} \sum_{i=1}^{K} R_{\text{p}}^{(i)}.
\end{equation}

\paragraph{Answer Reward.}
The answer-level reward is computed as
\begin{equation*}
    R_\text{a} =
    \begin{cases}
        1,   & \text{if } \mathrm{Judge}(\text{ref}, \mathbf{x}) = \mathbf{x} \succ \text{ref}, \\[2pt]
        0.5, & \text{if } \mathrm{Judge}(\text{ref}, \mathbf{x}) = \mathbf{x} \equiv \text{ref}, \\[2pt]
        0,   & \text{if } \mathrm{Judge}(\text{ref}, \mathbf{x}) = \mathbf{x} \prec \text{ref},
    \end{cases}
\end{equation*}
where ref is the high-quality reference response, and \(\mathrm{Judge}(\text{ref}, \mathbf{x})\) is the LLM-based evaluation function that compares \ the generated answer \(\mathbf{x}\) with \(\text{reference}\).

\section{Experiment}

\subsection{Experimental Setup}
\label{sec:exp_setting}
\paragraph{Training Details}
We first perform SFT on 3K examples using Qwen3 as the base model. Then, we conduct RL with Proximal Policy Optimization (PPO)~\cite{schulman2017proximal} starting from the SFT checkpoints. For reward computation, we employ DeepSeek-V3.1 as the judge for both process rewards and pairwise comparison (answer reward). Further training details are provided in the Appendix~\ref{appendix:exp details}.

\paragraph{Evaluation Benchmarks}
To comprehensively evaluate open-ended writing capabilities, we select representative benchmarks spanning multiple domains: WritingBench~\cite{wu2025writingbench} and HelloBench~\cite{que2024hellobench} for creative and general writing, DeepResearch-Bench~\cite{futuresearch2025deepresearchbenchevaluating} and DeepResearch-Gym~\cite{coelho2025deepresearchgymfreetransparentreproducible} for professional deep-research writing, and DiscoX~\cite{zhao2025discox} for discourse-level Chinese-English translation. Additional results on HelloBench and Deepresearch Bench are presented in Appendix~\ref{appendix:Experimental Results on Other Benchmarks} due to space limitations.
We follow the official evaluation protocols for WritingBench, HelloBench, and DiscoX. For DeepResearch-Gym, we use officially retrieved clueweb22~\cite{coelho2025deepresearchgymfreetransparentreproducible} materials as context and evaluate report relevance (Key Point Recall and Key Point Contradiction) and report quality.  More detailed descriptions are provided in the Appendix~\ref{appendix: evaluation_details}.
\vspace{-2mm}
\paragraph{Baselines.}
We select representative writing methods as baselines, including SFT-based methods such as Suri~\cite{pham-etal-2024-suri}, LongWriter~\cite{bai2024longwriter}, and Reverse-Engineering~\cite{wang2025reverse}, and RL-based methods such as Writing-RL~\cite{lei2025writing} and LongWriter-Zero~\cite{wu2025longwriter}. We implement all baselines using Qwen3-4B and Qwen3-8B as base models, with thinking mode enabled for fair comparison. More detailed descriptions are provided in Appendix~\ref{appendix: evaluation_details}.

\subsection{Main Results}
\vspace{-1mm}
\paragraph{Superior Performance Across Baseline Methods.}
\vspace{-1mm}
Table~\ref{tab:main results} shows that R\textsuperscript{2}-Write substantially outperforms both the base model and existing baselines across SFT-based (LongWriter, Reverse-Engineering) and RL-based (Writing-RL, LongWriter-Zero) methods. The integration of process reward supervision further enhances performance. 
These consistent improvements demonstrate that explicitly modeling reflection and revision patterns enables more effective deep reasoning for open-ended writing tasks.
\vspace{-2mm}
\paragraph{Generalization Across writing tasks.}
As shown in Table~\ref{tab:main results},  beyond the creative writing benchmark WritingBench, our method achieves significant improvements on research writing task DeepResearch-Gym and the expert-level translation task DiscoX. 
 On the one hand, this strong generalization stems from the diversity of R\textsuperscript{2}-Write's training data, which covers various writing scenarios. On the other hand, it also demonstrates that the reflection and revision patterns effectively generalize across diverse writing tasks.
These results confirm that reflection and revision are fundamental mechanisms for enhancing writing quality.

\subsection{Ablation Study}
\vspace{-1mm}
We conduct ablation studies to validate two key points: the effectiveness of reflection-revision patterns and the role of process reward supervision in RL training.
\subsubsection{Impact of Reflection-Revision Patterns}
Although our R\textsuperscript{2}-Write achieves state-of-the-art performance, the improvements may stem from two factors: knowledge distillation from the teacher model and higher-quality answers obtained through multi-turn iterative refinement. To verify that the performance gains primarily result from the incorporation of reflection and revision patterns, we conduct two ablation experiments: (1) \textbf{Qwen3-8B + Distill}: directly distilling from the Qwen3-30B-A3B-Thinking model, and (2) \textbf{Qwen3-8B + R\textsuperscript{2}-Write-Last}: distilling only the final answers from R\textsuperscript{2}-Write data while removing all intermediate thinking trajectories.

As shown in Table~\ref{tab:main results}, our \textbf{R\textsuperscript{2}-Write-SFT} outperforms both ablation baselines across all benchmarks. Compared to \textbf{Qwen3-8B + Distill}, our method demonstrates that explicit incorporation of reflection and revision patterns enhances deep reasoning capabilities beyond simple distillation from a teacher model, highlighting the advantages of our constructed thinking (enriched with reflection and revision patterns). More notably, \textbf{R\textsuperscript{2}-Write-SFT} achieves an average improvement of 10.1\% over \textbf{R\textsuperscript{2}-Write-Last}, confirming that our approach substantially enhances the effectiveness of deep reasoning for open-ended writing tasks.

\begin{takeaway}
The incorporation of reflection and revision patterns outperforms significantly direct distillation from teacher models, enabling more effective deep reasoning for open-ended writing tasks.
\end{takeaway}

\subsubsection{Effectiveness of Process Reward }

To further validate the role of process supervision in RL, we explore two aspects: (1) the quality of thinking trajectories. (2) The token count of thinking trajectories. 

To assess trajectory quality, we introduce \textbf{ProcessBench}, which evaluates reflection and revision segments from model outputs following Section~\ref{sec:rl}. Each segment is evaluated on three criteria: the validity of the identified issue, the alignment of the revision suggestion with rubrics, and whether the final answer implements the proposed revision. This allows us to evaluate the thinking process quality of any model on a given benchmark. Detailed prompts and implementation are provided in the Appendix~\ref{appendix: processbench}.

\begin{figure}[t!]
  \centering
  \vspace{-5pt}
  \resizebox{0.5\textwidth}{!}{
  \includegraphics[width=\textwidth]{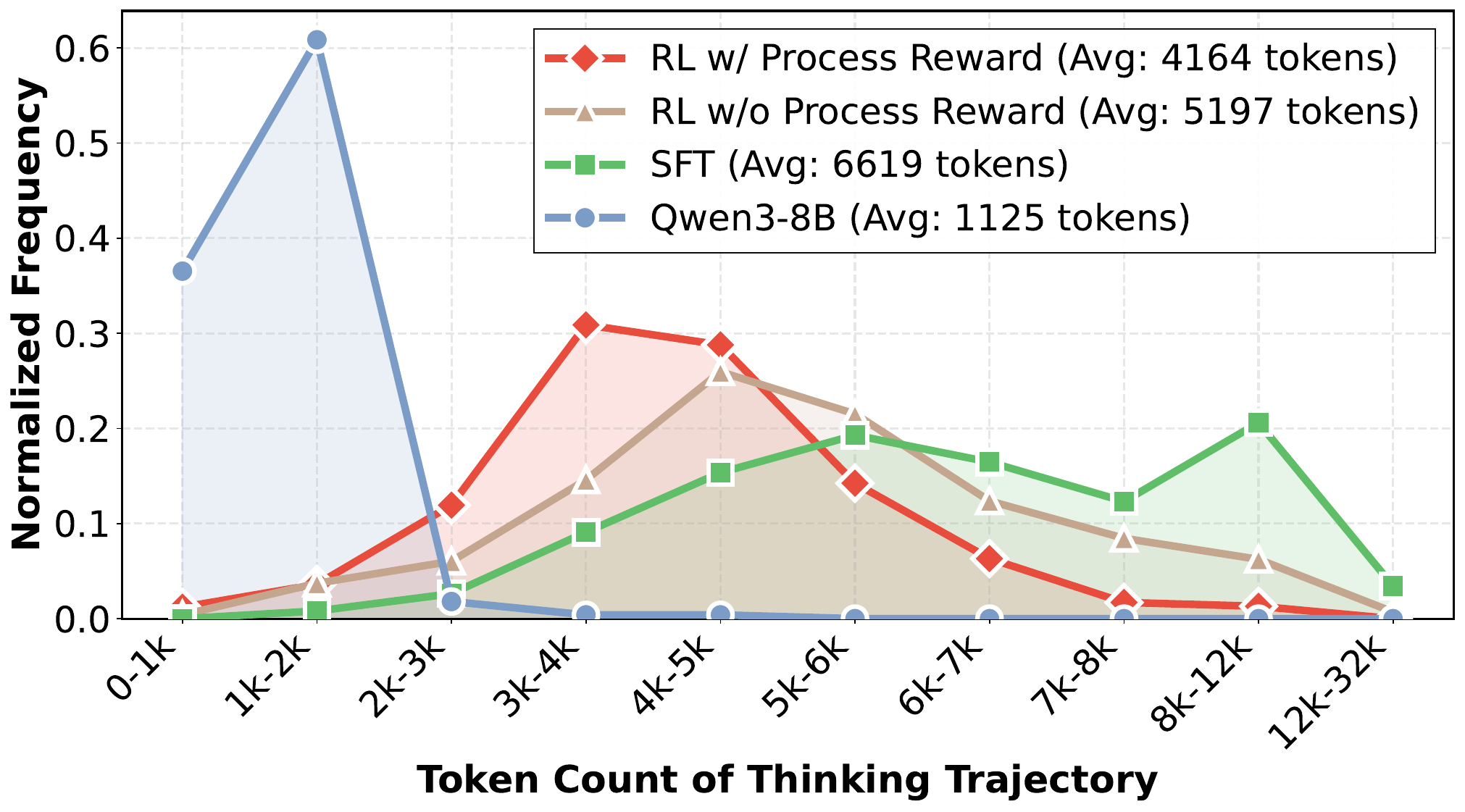}
  }
\caption{
Token length distribution of thinking trajectories across different methods on Writing Bench.
}
\label{fig:token_count}
\end{figure}

Table~\ref{tab:process_quality} shows that process reward supervision substantially improves ProcessBench scores, demonstrating enhanced reflection and revision quality. Additionally, Figure~\ref{fig:token_count} reveals that process reward effectively reduces thinking trajectory lengths by 20\% while maintaining performance gains compared to standard RL.
Overall, our process reward mechanism  achieves dual benefits: improved writing quality and more efficient reasoning through concise, high-quality reflection and revision.

\begin{table}[h]
\centering
\caption{\textbf{ProcessBench Results} on WritingBench and DeepResearch Gym. We use Qwen3-8B as the base model. ``RL'' denotes the baseline RL setup consistent with the main experiments, and ``$\text{RL}_\text{p}$'' denotes RL with process reward supervision.}
\vskip 0.05in
\resizebox{\columnwidth}{!}{
\begin{tabular}{lcc}
\toprule
\textbf{Method} & \textbf{WritingBench} & \textbf{DeepResearch Gym} \\
\midrule
R\textsuperscript{2}-Write + SFT & 0.586 & 0.541 \\
R\textsuperscript{2}-Write + SFT + RL & 0.660 & 0.584 \\
R\textsuperscript{2}-Write + SFT + $\text{RL}_\text{p}$ & \textbf{0.805} & \textbf{0.729} \\
\bottomrule
\end{tabular}
}

\label{tab:process_quality}
\end{table}


\section{Analysis}
In this section, we further investigate two key questions: (1) \textit{How do reflection and revision patterns enhance writing performance?} Specifically, what types of writing issues do these patterns help resolve? (2) \textit{Does R\textsuperscript{2}-Write affect the model's general capabilities beyond writing?}

\subsection{Q1: How R\textsuperscript{2}-Write Enhances Writing Performance.}

To investigate how the introduced reflection and revision patterns improve writing capabilities, we analyze specific cases where our method outperforms the baseline. Specifically, we compare \textit{Qwen3-8B + R\textsuperscript{2}-Write-SFT + $\text{RL}_\text{p}$} against \textit{Qwen3-8B + Distill + RL} on WritingBench and DeepResearch Gym. We analyze cases where R\textsuperscript{2}-Write achieves higher scores and manually categorize the scenarios into three types: (1) \textbf{Requirement Alignment (RA)}: correcting outputs that fail to match explicit user requirements, such as missing requested elements or violating format constraints; (2) \textbf{Factual \& Logical Correction (FLC)}: fixing errors in facts, data, or reasoning, including incorrect citations or logical contradictions; and (3) \textbf{Quality Enhancement (QE)}: improving overall writing quality through enhanced completeness, clarity, coherence, and depth.

We then use DeepSeek-V3.1 to extract segments containing reflection and revision patterns, and classify those segments that genuinely contribute to improving response quality (The details are provided in the Appendix~\ref{sec: Categorizing How Reflection and Revision Patterns Improve Writing Performance}). The classification results are shown in Table~\ref{tab:revision_distribution}. We find that the majority of improvements stem from Quality Enhancement, which aligns with the rubrics we designed for data construction. Additionally, we find that DeepResearch Gym tasks exhibit a higher proportion of Factual \& Logical Correction, where the model corrects contradictions between response statements and provided materials.

\begin{table}[t]
\centering
\caption{Distribution of revision patterns across different datasets. The table shows the percentage of each revision type (RA: Requirement Alignment, FLC: Factual \& Logical Correction, QE: Quality Enhancement) identified in the reflection and revision process for WritingBench and DeepResearch Gym datasets.\label{tab:revision_distribution}}
\vskip 0.05in
\resizebox{0.45\textwidth}{!}{

\begin{tabular}{lccc}
\toprule
\textbf{Dataset} & \textbf{RA} & \textbf{FLC} & \textbf{QE} \\
\midrule
WritingBench & 10.50\% & 19.88\% & 69.62\% \\
DeepResearch Gym & 24.17\% & 34.72\% & 41.11\% \\
\bottomrule
\end{tabular}}

\end{table}

\paragraph{Case Study.}
We provide concrete examples in the Appendix~\ref{sec:case study}, further illustrating how reflection and revision patterns enhance writing performance.

\begin{takeaway}
R\textsuperscript{2}-Write enables models to learn reflection and revision patterns based on rubrics that embody human writing standards, as systematically defined in our training framework, proving crucial for open-ended writing tasks.
\end{takeaway}

\subsection{Q2: Impact on General Capabilities}
To verify whether introducing reflection and revision patterns into the thinking chain affects the model's performance on non-writing tasks, we evaluate R\textsuperscript{2}-Write on mathematical reasoning and general knowledge benchmarks. Specifically, we compare R\textsuperscript{2}-Write-SFT against standard SFT that distills Qwen3-30B-A3B on the same queries, and R\textsuperscript{2}-Write-SFT + $\text{RL}_\text{p}$ against standard RL that uses the standard SFT checkpoint as initialization. For mathematical reasoning tasks, we evaluate on AIME25. For general knowledge tasks, we evaluate on MMLU-Pro~\cite{wang2024mmlu}.

 The results in Table~\ref{tab:non_writing_performance} show that R\textsuperscript{2}-Write maintains comparable performance across other tasks, demonstrating that our method does not compromise the model's general capabilities while improving its writing capability.

 \begin{table}[t]
\centering
\caption{Performance comparison on non-writing tasks. R\textsuperscript{2}-Write maintains competitive performance on AIME25 and MMLU-Pro, demonstrating that reflection and revision patterns do not compromise the model's general capabilities.}
\vskip 0.05in
\resizebox{0.45\textwidth}{!}{
\begin{tabular}{lcc}
\toprule
\textbf{Method} & \textbf{AIME25} & \textbf{MMLU-Pro} \\
\midrule
Qwen3-8B + Distill & 66.78 & 59.62 \\
R\textsuperscript{2}-Write-SFT & 67.84 & 60.42 \\
\midrule
Qwen3-8B + Distill + RL & 65.50 & 58.28 \\
R\textsuperscript{2}-Write-SFT + $\text{RL}_\text{p}$ & 64.78 & 59.22 \\
\bottomrule
\end{tabular}
}

\label{tab:non_writing_performance}
\end{table}

\begin{table}[t]
\centering
\caption{Ablation study on reward design and RL algorithms. All results are evaluated on WritingBench.}
\vskip 0.05in
\resizebox{0.45\textwidth}{!}{
\begin{tabular}{lcc}
\toprule
\textbf{Method} & \textbf{Reward $\times$ Algorithm} & \textbf{Score} \\
\midrule
Qwen3-8B (w/o RL) & -- & 71.84 \\
\midrule
\multicolumn{3}{l}{\textit{Pointwise Reward:}} \\
\quad + PPO & Pointwise $\times$ PPO & 73.42 \\
\quad + GRPO & Pointwise $\times$ GRPO & 75.20 \\
\midrule
\multicolumn{3}{l}{\textit{Pairwise Reward:}} \\
\quad + GRPO & Pairwise $\times$ GRPO & 77.87 \\
\quad + PPO & Pairwise $\times$ PPO & \textbf{78.42} \\
\bottomrule
\end{tabular}}

\label{tab:reward_rl_comparison}
\end{table}

\subsection{Analysis of RL Design Variants}

\paragraph{Analysis of Reward Design.}
To provide effective rewards, we use pairwise comparison with high-quality references. Here, we compare the pairwise reward against the widely-adopted pointwise grading method~\citep{zheng2023judging, liu2025inference}, which utilizes a judge LLM to provide a scalar rating representing response quality. We follow the experiment setting in Section~\ref{sec:exp_setting}. As shown in Table~\ref{tab:reward_rl_comparison}, the pairwise reward mechanism with reference-based comparison demonstrates superiority in providing more discriminative rewards, effectively incentivizing the model to advance writing capabilities.

\paragraph{Analysis of RL Algorithms.}
We also compare PPO against the commonly-used GRPO algorithm~\cite{shao2024deepseekmath} to assess their effectiveness for training reflection and revision patterns. As shown in Table~\ref{tab:reward_rl_comparison}, PPO consistently outperforms GRPO with pointwise or pairwise rewards, demonstrating superior performance for writing tasks.

\section{Conclusion}
In this paper, we reveal that current reasoning models severely underutilize reflection and revision patterns, resulting in limited performance gains compared to mathematical reasoning. To address this gap, we propose \text{R\textsuperscript{2}-Write}, an automated framework that enriches thinking trajectories with effective reflection and revision patterns through writer-judge collaboration and process reward supervision. Extensive experiments validate that explicit incorporation of these patterns substantially enhances writing performance while maintaining token efficiency. Our work establishes a foundation for applying test-time scaling to open-ended domains and highlights the critical importance of task-specific reasoning pattern design.



\section*{Impact Statement}

This paper presents work whose goal is to advance the field of 
Machine Learning. There are many potential societal consequences 
of our work, none which we feel must be specifically highlighted here.

\nocite{langley00}

\bibliography{example_paper}
\bibliographystyle{icml2025}

\newpage
\appendix
\onecolumn

\section{Related Works}

\subsection{Open-ended Writing}
 \paragraph{Open-ended Writing Benchmarks.}
Early work on open-primarily focused on long-form text generation capabilities. For instance, LongBench-Write~\cite{bai2024longbench} evaluates ultra-long-form text generation (>10,000 words). HelloBench~\cite{que2024hellobench} evaluates performance on diverse "in-the-wild" tasks sourced from real user queries, gauging practical applicability across various writing scenarios.
Recently, there has been a shift toward rubric-based evaluation frameworks that provide more fine-grained assessment. WritingBench~\cite{wu2025writingbench} and DiscoX~\cite{zhao2025discox} construct detailed checklists based on task-specific rubrics, enabling LLM judges to perform structured evaluation. Similarly, some deepresearch benchmarks~\cite{futuresearch2025deepresearchbenchevaluating, coelho2025deepresearchgymfreetransparentreproducible, wang2025liveresearchbench, wan2025deepresearch,fan2025understanding, zheng2025deepresearcher, java2025characterizing, liang2025towards} establish multi-dimensional rubrics covering relevance, comprehensiveness, depth, and fluency to assess report quality from multiple perspectives. These benchmarks enable more objective and consistent evaluation of open-ended writing quality.

 \paragraph{Open-ended Writing Methods.}
Early work focused on constructing long-generation post-training datasets for fine-tuning~\cite{bai2024longbench, pham-etal-2024-suri, tu2025longwriter, quan2024language}. Some approaches distill high-quality responses from teacher models and synthesize reasoning trajectories in reverse~\cite{wang2025reverse}.
Recently, RL-based methods have emerged for open-ended writing. LongWriter-Zero~\cite{wu2025longwriter} employs RL from scratch without annotated data to develop ultra-long text generation capabilities. Writing-RL~\cite{lei2025writing} proposes an Adaptive Curriculum Reinforcement Learning framework to advance long-form writing beyond SFT. Deep-Zero~\cite{lu2025writing} introduces a writing-principle-based pairwise Generative Reward Model and Bootstrapped Relative Policy Optimization algorithm for dynamic, reference-free training. However, the role of deep reasoning for open-ended writing tasks remains largely unexplored, hindering further improvement.

\subsection{Deep Reasoning}

The test-time scaling paradigm has driven substantial breakthroughs in verifiable domains through reinforcement learning with verifiable rewards (RLVR)~\cite{guo2025deepseek, zhang2025survey, wen2025reinforcement}. Deep reasoning models achieve remarkable improvements in mathematics~\cite{liu2025qfft, guan2025rstar, yu2025z1}, where clear reward signals enable effective multi-step reasoning and self-correction.
Recent work extends deep reasoning to scientific and general tasks~\cite{huan2025does, chen2025reinforcement, yu2025rlpr, wen2025reinforcement}. However, its effectiveness for open-ended writing—where no ground-truth exists—remains largely unexplored. This work investigates how deep reasoning benefits open-ended writing and proposes methods to effectively incorporate reasoning patterns into this domain.

\section{Details of Training Dataset Construction}

\subsection{Query Section}
\label{appendix: Query Section}
To enhance the generalization capability of our trained model, we ensure data diversity by curating a dataset that encompasses multiple types of open-ended writing tasks. Specifically, we include both creative-writing queries and report-generation tasks, where the latter corresponds to deep-research writing scenarios in which users provide specific requirements along with collected materials to generate structured, comprehensive reports.

\paragraph{Creative Writing Task.}
For creative writing, we use the \textsc{DeepWriting} subset (20K instances) from~\cite{wang2025reverse}, a high-quality dataset covering over 14 categories. We filter each category to remove excessively short queries (shorter than 10 tokens) and eliminate semantically similar queries using n-gram overlap: we compute the Jaccard similarity of 3-grams between query pairs and remove queries with similarity scores above 0.7.

\begin{algorithm}[t]
\small
\caption{Writer–Judge Data Synthesis with Reflection and Revision}
\label{alg:actor_critic}
\KwIn{Writer Model $\mathcal{A}$, Judge Model $\mathcal{B}$, dataset 
$\mathcal{D} = \{(q, \rho_q)\}$, max turns $T_{\max}$, target score $S_{\text{target}}$}
\KwOut{Training trajectories with thinking process and final answer}

\ForEach{$(q, \rho_q) \in \mathcal{D}$}{
  \tcp{Step 1: Initial thinking and answer}
  $\tau_q \leftarrow [\,]$; 
  $T_0, a_q^{(0)} \leftarrow \mathcal{A}(q)$\;
  
  \tcp{Step 2: Initial evaluation (get score $s_q^{(0)}$ and feedback $f_q^{(0)}$)}
  $(s_q^{(0)}, f_q^{(0)}) \leftarrow \mathcal{B}(a_q^{(0)}, q, \rho_q)$;
  append $[T_0, a_q^{(0)}]$ to $\tau_q$\;
  
  $t \leftarrow 1$\;
  \While{$t < T_{\max}$ \textbf{and} $s_q^{(t)} < S_{\text{target}}$}{
    \tcp{Step 3: Reflection and Revision Segment $R_q^{(t)}$}
    $R_q^{(t)} \leftarrow \mathcal{A}(q, a_q^{(t-1)}, f_q^{(t-1)})$\;
    
    \tcp{Step 4: Revised answer}
    $a_q^{(t)} \leftarrow \mathcal{A}(q, a_q^{(t-1)}, R_q^{(t)})$\;
    
    \tcp{Step 5: Re-evaluation}
    $(s_q^{(t)}, f_q^{(t)}) \leftarrow \mathcal{B}(a_q^{(t)}, q, \rho_q)$\;
    
    \If{$s_q^{(t)} > s_q^{(t-1)}$}{
      append $[R_q^{(t)}, a_q^{(t)}]$ to $\tau_q$\;
    }
    \Else{
      $(s_q^{(t)}, f_q^{(t)}) \leftarrow (s_q^{(t-1)}, f_q^{(t-1)})$ \tcp*{keep previous score}
    }
    $t \leftarrow t + 1$\;
  }
  
  \tcp{Step 6: Assemble trajectory}
  $k \leftarrow t$; 
  $a_q^\star \leftarrow a_q^{(k)}$\;
  
  $\text{Think}_q \leftarrow T_0 \oplus a_q^{(0)} \oplus \bigoplus_{i=1}^{k}[R_q^{(i)}]$\;
  
  $\text{Sample}_q \leftarrow \texttt{<think>}\;\text{Think}_q\;\texttt{</think>}\;\texttt{<answer>}\;a_q^\star\texttt{</answer>}$\;
}
\Return{$\{\text{Sample}_q\}_{(q,\rho_q)\in\mathcal{D}}$}\;
\end{algorithm}

\paragraph{Report Generation Task.}
For report generation, we collect 10K real-world user deep-research queries from public online websites for DeepResearch, spanning 10 major categories. We apply the same filtering strategy to each category, removing excessively short queries (shorter than 10 tokens) and highly similar queries based on n-gram overlap. For each query, we retrieve materials using Google Search and remove low-quality text from each material to serve as supplementary context. We then concatenate the retrieved materials with the query to form the final query. We truncate queries to ensure each does not exceed 10K tokens.

For both creative writing and report generation queries, we identify challenging queries with substantial performance gaps, where the model most requires reflection and revision. For each query $q$, we generate a response using Qwen3-30B-A3B and compute rubric-based scores. Let $S^q_r$ and $S^q_m$ denote the aggregated scores of the reference answer and the model's answer, respectively, combining both query-specific and generic evaluation dimensions. We define the performance gap as:
\[
\Delta_q = S^q_r - S^q_m,
\]
and select the top-$k$ queries with the largest gaps. This difficulty-based filtering yields 3K high-quality instances for SFT (in \S~\ref{sec:3.2}) and 5K instances for RL (in \S~\ref{sec:rl}), ensuring our training focuses on genuinely challenging cases that maximize model improvement.

\begin{figure*}[t]
    \centering
    \begin{subfigure}[b]{0.48\textwidth}
        \centering
        \includegraphics[width=\textwidth]{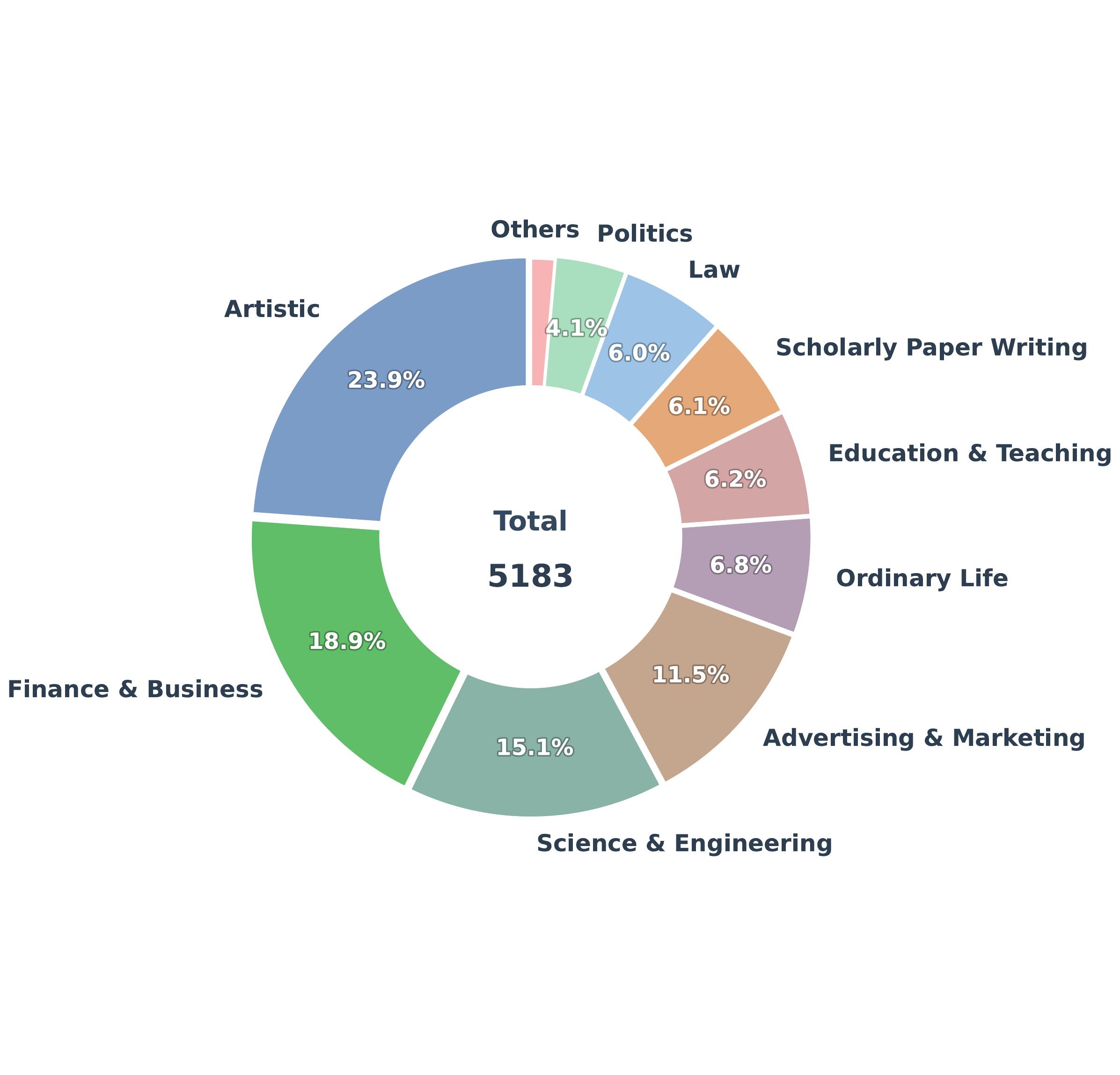}
        \caption{Creative Writing}
        \label{fig:creative_writing_dist}
    \end{subfigure}
    \hfill
    \begin{subfigure}[b]{0.49\textwidth}
        \centering
        \includegraphics[width=\textwidth]{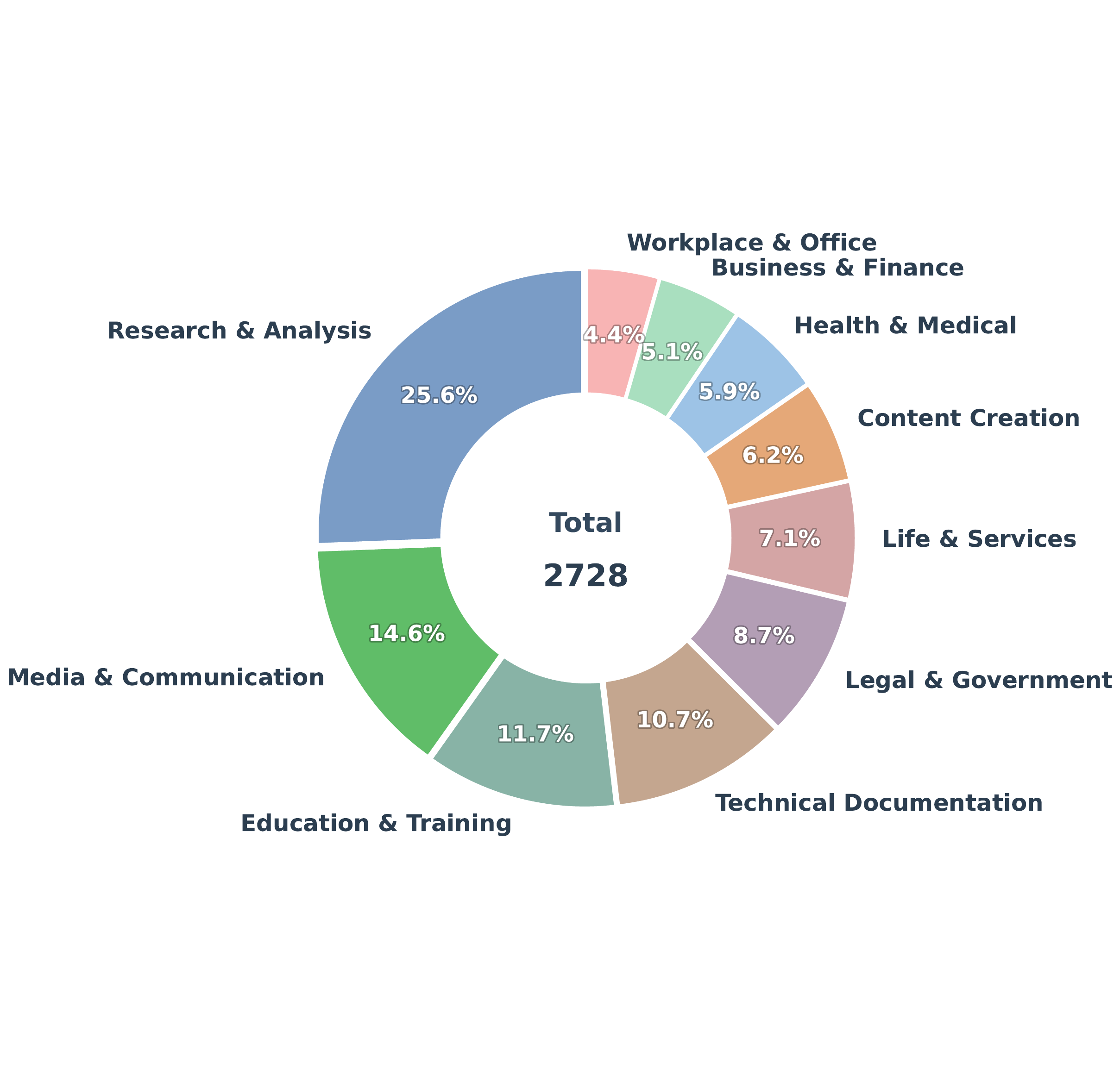}
        \caption{Report Generating}
        \label{fig:report_generating_dist}
    \end{subfigure}
    \caption{Data distribution of our constructed training set, which includes both SFT and RL data. (a) shows the domain distribution for creative writing tasks, and (b) shows the category distribution for report generating tasks.}
    \label{fig:data_distribution}
\end{figure*}

\subsection{Evaluation Rubrics Construction}

\label{appendix: Evaluation Rubrics Construction}

Since open-ended writing tasks lack ground-truth answers, we design evaluation rubrics that align with human writing standards.
Following prior work~\cite{coelho2025deepresearchgymfreetransparentreproducible, futuresearch2025deepresearchbenchevaluating}, we construct two types of rubrics using Claude-4.5-Sonnet for each query: \emph{query-specific rubrics} and \emph{general quality rubrics}. Query-specific rubrics define fine-grained, task-dependent criteria focusing on coverage of key information required by the prompt (Figure~\ref{fig:rubric_generation}). General quality rubrics assess overall answer quality, including fluency, completeness, and creativity (Figure~\ref{fig:adaptive_criteria}). For report-generation tasks, we additionally incorporate a factual consistency check that verifies whether report statements align with the provided materials.

\subsection{Rubrics Quality Validation}
\label{sec:human_alignment}

To validate the quality of our LLM-based evaluation rubrics, we conducted a systematic alignment study on 100 randomly sampled cases (50 creative writing, 50 report generation) from our constructed training dataset.

\textbf{Experimental Setup.} 
For each case, we generated responses using Qwen3-30B-A3B and collected rubric-based evaluations from three LLM judges: GPT-5, Gemini-2.5-Pro, and DeepSeek-V3.1. Each judge provided scores on two rubric dimensions: query-specific rubrics (normalized to 0-1 scale) and general quality rubrics (normalized to 0-10 scale).

\textbf{Inter-Judge Consistency.}
Table~\ref{tab:llm_scores} shows the average scores assigned by each judge. All three judges demonstrated relatively consistent scoring patterns with small score variations, indicating reliable automated evaluation across different models.

\begin{table}[h]
\centering
\caption{Average scores by LLM judges. Query-specific rubrics are on 0-1 scale, general quality rubrics on 0-10 scale.}
\label{tab:llm_scores}
\vskip 0.05in
\begin{tabular}{lcccc}
\toprule
\textbf{Rubric Type} & \textbf{GPT-5} & \textbf{Gemini-2.5-Pro} & \textbf{DeepSeek-V3.1} & \textbf{Std.} \\
\midrule
Query-Specific (0-1) & 0.82 & 0.89 & 0.81 & 0.04 \\
General Quality (0-10) & 7.35 & 7.05 & 7.25 & 0.15 \\
\bottomrule
\end{tabular}
\end{table}

\textbf{Human Validation.}
To further ensure quality, we recruited 3 human experts to independently review each LLM judge's evaluation, determining whether they agreed with the scores and reasoning. An evaluation was marked as ``aligned'' if experts agreed with both the score and key justifications.
Table~\ref{tab:human_alignment} presents the alignment results. DeepSeek-V3.1 achieved the highest alignment (92.0\%), with an overall average of 90.3\% across all judges, demonstrating that LLM-based rubric evaluation reliably approximates human judgment.

\begin{table}[h]
\centering
\caption{Human-LLM alignment rates across task types.}
\label{tab:human_alignment}
\vskip 0.05in
\begin{tabular}{lcccc}
\toprule
\textbf{Task Type} & \textbf{GPT-5} & \textbf{Gemini-2.5-Pro} & \textbf{DeepSeek-V3.1} & \textbf{Avg.} \\
\midrule
Creative Writing & 89.0\% & 88.0\% & 91.0\% & 89.3\% \\
Report Generation & 91.0\% & 90.0\% & 93.0\% & 91.3\% \\
\midrule
\textbf{Overall} & \textbf{90.0\%} & \textbf{89.0\%} & \textbf{92.0\%} & \textbf{90.3\%} \\
\bottomrule
\end{tabular}
\end{table}

\paragraph{Reference Data Construction}
Since our RL training requires high-quality references for pairwise comparison, we construct references as follows. For creative writing data sourced from Reverse-Engineering~\cite{wang2025reverse}, we use their official references. For report generation data, we generate responses using Claude-4.5-Sonnet, Gemini-2.5-Pro, and GPT-5, and select the highest-scoring response based on our rubrics as the final reference.
 
\subsection{Data Construction Pipeline}
\label{sec: Data Construction Pipeline}
Our data construction pipeline employs writer-judge interactions to synthesize reflection-enriched thinking trajectories. The judge model evaluates the writer's output and provides detailed feedback based on task-specific rubrics. The writer then performs self-refinement in two steps: (1) \textbf{Reflection generation.} The writer rewrites the judge's feedback into its own internal reasoning, explicitly identifying issues in the current answer (\emph{reflection}) and proposing concrete fixes (\emph{revision}). We inject human-like thinking patterns by encouraging exploratory phrases such as "Hmm...maybe I should revise..." or "Wait, I found that...", creating natural reflection while avoiding rigid, formulaic thinking. The prompt is shown in Figure~\ref{fig:self_revision_1}. (2) \textbf{Answer rewrite.} The writer generates a revised answer conditioned on the reflection (Figure~\ref{fig:article_revision}).
The detailed algorithm is presented in Algorithm~\ref{alg:actor_critic}.

\paragraph{Hyperparameter Configuration.}
During evaluation, we compute both query-specific and general quality scores. 
For query-specific rubrics, scores are averaged across all key points with a maximum score of $1.0$; 
for general quality rubrics, scores are averaged across quality dimensions with a maximum score of $10.0$. 
We define threshold target scores $S_{\text{target}}$ as the minimum acceptable quality boundaries, 
setting $S_{\text{target}} = 1.0$ for query-specific evaluation and $S_{\text{target}} = 8.0$ 
for general quality assessment. Specifically, the refinement process terminates once the evaluated 
score meets or exceeds the corresponding threshold, \emph{i.e.}, $\text{Score} \geq S_{\text{target}}$. 
The maximum number of refinement iterations is set to $T_{\text{max}} = 3$, as empirical observations 
indicate that the majority of cases achieve satisfactory quality within $2$--$3$ rounds. 
This configuration strikes a balance between refinement effectiveness and computational efficiency, 
preventing over-refinement and mitigating excessive token consumption.

\subsection{Dataset Statistics}
Figure~\ref{fig:data_distribution} presents the distribution of our training dataset (SFT + RL data). 
The creative writing subset (Figure~\ref{fig:creative_writing_dist}) spans multiple domains, with Artistic content being predominant, followed by Finance, Science, and Education, along with other domains covering diverse creative scenarios.
The report generating subset (Figure~\ref{fig:report_generating_dist}) exhibits a balanced distribution across professional categories, including Research \& Analysis, Media \& Communication, Education \& Training, Technical Documentation, and other professional writing domains.

\section{Experimental Details}
\label{appendix:exp details}
\subsection{Details of Pilot Study}
\paragraph{Evaluation Details and Prompts}
\label{appendix: 2.1}
For math reasoning evaluation, we use the prompt shown in Figure~\ref{fig:prompt_eval_math}, performing 16 inference runs and computing the average score (pass@16). The sampling temperature is set to 0.6 with a maximum length of 32k tokens. For writing task evaluation, we use the prompt shown in Figure~\ref{fig:prompt_eval_writing}. The sampling temperature is set to 0.6 with a maximum length of 32k tokens.

\paragraph{Prompts of Classify Thinking Patterns}
\label{appendix: 2.2}

After obtaining the thinking trajectories from evaluation, we categorize the thinking patterns for math reasoning tasks including MATH500 and AIME25, as well as writing tasks including WritingBench and HelloBench. The classification prompt is shown in Figure~\ref{fig:pattern_classification}. For writing tasks, we replace the \texttt{Ground Truth} in the prompt with \texttt{Evaluation Rubrics}. We first calculate the proportion of each pattern type in the thinking trajectories, then analyze their contribution to producing correct answers.

\subsection{RL Training Details}

We conduct RL training using the VeRL framework~\cite{sheng2025hybridflow} with Proximal Policy Optimization (PPO)~\cite{schulman2017proximal} and Generalized Advantage Estimation (GAE) as the advantage estimator. We adopt an LLM-as-a-Judge paradigm where DeepSeek-V3.1 serves as the reward model, computing rewards via pairwise comparison between model outputs and high-quality references. The reward comprises process-level evaluation and answer-level evaluation.

\paragraph{Training Configuration.}
We train for 400 steps with a training batch size of 32. The maximum prompt length is set to 4,096 tokens and response length to 10,000 tokens to accommodate long-form generation. Rollout generation uses temperature 1.0, implemented via the vLLM engine with tensor model parallel size 2. We employ Fully Sharded Data Parallel (FSDP) with parameter/optimizer offloading for efficient multi-GPU training on 8$\times$A100 80GB GPUs. The writer model uses learning rate $1\times10^{-6}$ with warm-up ratio 0.4, while the critic uses learning rate $1\times10^{-5}$ with warm-up ratio 0.05. The KL divergence penalty coefficient is set to 0.001. We evaluate checkpoints on the validation set every 50 steps.

\paragraph{Reward Computation.}
For reward computation, we first calculate the answer-level reward using the prompt in Figure~\ref{fig:answer_reward}. We then sequentially apply the prompts in Figures~\ref{fig:process_reward_step1} and~\ref{fig:process_reward_step2} to extract reflection and revision segments and evaluate their quality along three dimensions: issue validity, revision quality, and implementation. Finally, we aggregate individual segment scores to obtain the overall process reward $R_\text{process}$ and combine it with the answer reward $R_\text{answer}$ to produce the total reward signal: $R_\text{total} = \alpha * R_\text{answer} + (1 - \alpha) \cdot R_\text{process}$, where $\alpha$  is set 0.25, which controls the relative weight of process supervision.

\begin{table*}[t]
    \centering
    
    \caption{Definitions and examples of the five reasoning patterns.}
    \small
    \begin{tabular}{p{3.0cm} | p{6.8cm} | p{5.0cm}}
        \toprule
        \centering\textbf{Pattern Type} & \centering\textbf{Definition} & \quad \quad  \quad  \quad  \quad \quad \textbf{Example} \\
        \midrule
        \vspace{0.2em}\textbf{Answer Verification}\vspace{0.2em}
        & The reasoning includes explicit or implicit steps that check intermediate computations or final results for correctness.
        & ``Let's verify this result by \dots'', ``Reviewing: the story is in the first person and within 500 words \dots'' \\
        \midrule
        \vspace{0.2em}\textbf{Backtracking}\vspace{0.2em}
        & The reasoning identifies an error or dead end and switches to a different approach or recalculates.
        & ``I forgot to address the second question; I need to rewrite the ending \dots'', ``Wait, I made an error. Let me recalculate \dots'' \\
        \midrule
        \vspace{0.2em}\textbf{Subgoal Setting}\vspace{0.2em}
        & The reasoning explicitly introduces intermediate subgoals that decompose the problem into smaller parts.
        & ``First, I'll calculate \dots then I'll \dots'', ``Step 1: Find the area. Step 2: Calculate the perimeter.'' \\
        \midrule
        \vspace{0.2em}\textbf{Backward Chaining}\vspace{0.2em}
        & The reasoning starts from the target result and works backward to infer the required inputs or steps.
        & ``To get $24$, I need $24 \div 2 = 12$ \dots'', ``Working backwards from answer \dots'' \\
        \midrule
        \vspace{0.2em}\textbf{Summarization}\vspace{0.2em}
        & The reasoning summarizes completed subtasks or current progress and indicates the remaining steps.
        & ``Now we have obtained $x = 5$, next we need to \dots'', ``So far we've found \dots the remaining step is \dots'' \\
        \bottomrule
    \end{tabular}
    
    \label{tab:define_reasoning_patterns}
\end{table*}

\subsection{Evaluation Benchmarks}
\label{appendix: evaluation_details}
We evaluate our approach on five representative benchmarks covering diverse open-ended writing scenarios:

\begin{itemize}
    \item \textbf{WritingBench}~\cite{wu2025writingbench}: A comprehensive benchmark with over 1,200 tasks across six writing domains (creative, persuasive, informative, technical, business, and legal), spanning 100 subdomains. Each task is evaluated on five dynamic criteria including coherence and relevance. We use the fine-tuned critic model in the original paper to evaluate the responses. 
    
    \item \textbf{DeepResearchBench}~\cite{futuresearch2025deepresearchbenchevaluating}: A benchmark of 100 PhD-level research tasks across 22 fields, designed to evaluate deep research agents' capabilities in producing analyst-grade, citation-rich reports. Evaluation employs reference-based adaptive criteria for report quality assessment and citation-based metrics for information retrieval capabilities. We use the model Claude-3.7-Sonnet for evaluation.
    
    \item \textbf{DeepResearchGym}~\cite{coelho2025deepresearchgymfreetransparentreproducible}: An open-source sandbox combining a reproducible search API with rigorous evaluation protocols. It extends the Researchy Questions benchmark with LLM-as-a-judge metrics measuring alignment with information needs, retrieval faithfulness, and report quality across large-scale web corpora (ClueWeb22 and FineWeb). We use the model Claude-3.7-Sonnet for evaluation.
    
    \item \textbf{HelloBench}~\cite{que2024hellobench}: A hierarchical benchmark evaluating long text generation across five task types based on Bloom's Taxonomy: open-ended QA, summarization, chat, text completion, and heuristic generation. It employs HelloEval, a human-aligned evaluation method that reduces evaluation time while maintaining high correlation with human judgments. We use the model Claude-3.7-Sonnet for evaluation.
    
    \item \textbf{DiscoX}~\cite{zhao2025discox}: A discourse-level translation benchmark with 200 professionally-curated texts from 7 expert domains, averaging over 1,700 tokens. Evaluation uses Metric-S, a reference-free system providing fine-grained assessments across accuracy, fluency, and appropriateness with strong human judgment alignment. We use the model Claude-3.7-Sonnet for evaluation.
    
\end{itemize}

For creative writing tasks, we strictly follow the evaluation protocols and prompts specified in each benchmark's original paper and conduct all evaluations ourselves. 
For deep research benchmarks, we provide retrieved materials as context and focus evaluation on writing capabilities. Specifically, for DeepResearch-Gym, we use officially retrieved ClueWeb22~\cite{coelho2025deepresearchgymfreetransparentreproducible} materials and evaluate report relevance (Key Point Recall and Key Point Contradiction) and report quality. For DeepResearch-Bench, we use materials retrieved by Tongyi DeepResearch 30B-A3B~\cite{team2025tongyi} and evaluate four dimensions: Comprehensiveness, Insight/Depth, Instruction-Following, and Readability.

For baseline comparisons, we re-evaluate open-source baseline models on these benchmarks using their official implementations. For detailed evaluation settings and prompts, please refer to the original papers.
\subsection{Details of Process Bench}
\label{appendix: processbench}
To further validate the role of process supervision in RL, we examine two aspects: (1) the token distribution of thinking trajectories, and (2) the quality of thinking trajectories. 

For quality evaluation, we introduce ProcessBench. Following Section~\ref{sec:rl}, we use an LLM to extract all reflection segments where the model explicitly identifies issues and proposes solutions: $\mathcal{M} = \{M_{F_1}, \dots, M_{F_K}\}$. Each segment $M_{F_i}$ is evaluated on three dimensions:
\begin{itemize}
    \item \textbf{Issue Identification} $R_{\text{find}}^{(i)} \in \{-1, +1\}$: valuable vs. spurious issue.
    \item \textbf{Revision Suggestion} $R_{\text{rev}}^{(i)} \in \{-1, +1\}$: revision aligns with vs. violates rubrics.
    \item \textbf{Execution Alignment} $R_{\text{align}}^{(i)} \in \{-1, +1\}$: final answer implements the revision or not.
\end{itemize}

For each segment, the process score is computed as:
\begin{equation}
    R_{\text{p}}^{(i)} = 
    \begin{cases}
        +1, & \text{if } R_{\text{find}}^{(i)} > 0 \text{ and } R_{\text{rev}}^{(i)} > 0 \text{ and } R_{\text{align}}^{(i)} > 0, \\
        -1, & \text{otherwise.}
    \end{cases}
\end{equation}

This allows us to evaluate the thinking process quality of any model on a given benchmark. 

\subsection{Categorizing How Reflection and Revision Patterns Improve Writing Performance}
\label{sec: Categorizing How Reflection and Revision Patterns Improve Writing Performance}
To investigate how reflection and revision patterns improve writing capabilities, we analyze cases where R\textsuperscript{2}-Write-SFT + $\text{RL}_\text{p}$ outperforms Qwen3-8B + Distill + RL on WritingBench and DeepResearchGym. We manually categorize improvement scenarios into three types: (1) \textbf{Requirement Alignment (RA)}: correcting outputs that fail to meet explicit user requirements, such as missing requested elements or violating format constraints; (2) \textbf{Factual \& Logical Correction (FLC)}: fixing errors in facts, data, or reasoning, including incorrect citations or logical contradictions; and (3) \textbf{Quality Enhancement (QE)}: improving overall writing quality through enhanced completeness, clarity, coherence, and depth.

Our categorization process involves two stages: first, we use DeepSeek-V3.1 to provide preliminary analysis of how patterns contribute to improvements; then, we employ Claude-4.5-Sonnet to cluster and refine these initial annotations into the final three-category taxonomy. Finally, we employ DeepSeek-V3.1 to quantify the prevalence of each improvement category (prompt in Figure~\ref{fig:revision_classification}).

The classification results are shown in Table~\ref{tab:revision_distribution}. We find that Quality Enhancement accounts for the majority of improvements, aligning with our rubric design during data construction. Notably, DeepResearchGym tasks exhibit a higher proportion of Factual \& Logical Correction, where the model identifies and corrects contradictions between response statements and provided materials.

\subsection{Experiment Cost}

We provide a comprehensive breakdown of computational costs across data construction, evaluation, and training stages. Table~\ref{tab:overall_cost} presents API costs for data construction (using Claude-4.5-Sonnet) and evaluation (using Claude-3.7-Sonnet), as well as GPU hours for training on NVIDIA A100 GPUs.

\begin{table}[h]
\centering
\caption{Overall Cost Breakdown}
\label{tab:overall_cost}
\resizebox{0.45\textwidth}{!}{
\begin{tabular}{@{}llr@{}}
\toprule
\textbf{Category} & \textbf{Item} & \textbf{Cost} \\
\midrule
\multirow{2}{*}{Dataset Construction} 
    & Rubrics Generation & \$81 \\
    & Pipeline (Judge model) & \$52 \\
\midrule
\multirow{2}{*}{Evaluation} 
    & WritingBench & \$1.9 \\
    & Deepresearch Gym & \$2.7 \\
\midrule
\multirow{2}{*}{Training (A100)} 
    & SFT Training & 40 GPU hrs \\
    & RL Training & 580 GPU hrs \\
\bottomrule
\end{tabular}}
\end{table}

\section{Additional Experiments}

\subsection{Experimental Results on Other Benchmarks}
\label{appendix:Experimental Results on Other Benchmarks}
We further evaluate our approach on HelloBench and DeepResearchBench, with results presented in Table~\ref{tab:main results_appendix}. Our method consistently outperforms all SFT-based and RL-based baselines across these benchmarks, demonstrating strong generalization and effectiveness across diverse open-ended writing tasks.

\begin{table*}[t]
\centering
\caption{Performance comparison across different training methods on HelloBench and DeepresearchBench datasets. $\uparrow$ indicates higher is better.}
\resizebox{1\textwidth}{!}{
\begin{tabular}{lccccccc}
\toprule
\multirow{2}{*}{\textbf{Method}} & \multirow{2}{*}{\textbf{Training}} & \multicolumn{1}{c}{\textbf{HelloBench}} & \multicolumn{5}{c}{\textbf{DeepresearchBench}} \\
\cmidrule(lr){3-3} \cmidrule(lr){4-8}
& & Avg. Score($\uparrow$) & Overall($\uparrow$) & Comp.($\uparrow$) & Depth($\uparrow$) & Inst.($\uparrow$) & Read($\uparrow$) \\
\midrule

\multicolumn{7}{c}{\textbf{SFT-Based Methods}} \\
\midrule
Qwen3-4B & - & 65.86 & 38.11 & 36.32 & 33.41 & 43.66  & 42.33 \\
Qwen3-8B & - & 71.80 & 40.52 & 38.74 & 36.27 & 45.85  & 44.18 \\
Longwriter-9B & SFT & 62.00 & 38.68 & 36.84 & 34.45 & 43.18 & 40.26 \\
Reverse-Engineering-8B & SFT & 74.90 & 42.42 & 41.20 & 39.44 & 44.18 & 44.86 \\
 \rowcolor{lightgreen} Qwen3-4B + R\textsuperscript{2}-Write-SFT  & SFT & \text{76.36} & 44.18 & 42.31 & 41.89 & 47.56 & 47.06 \\
 \rowcolor{lightgreen} Qwen3-8B + R\textsuperscript{2}-Write-SFT  & SFT & \textbf{80.32} & \textbf{46.00} &  \textbf{43.32} & \textbf{42.86} & \textbf{48.75} & \textbf{49.08} \\
\midrule
\multicolumn{7}{c}{\textbf{RL-Based Methods}} \\
\midrule
Qwen3-4B + RL & PPO & 70.20 & 39.74 &  40.25 & 37.80 & 41.22 & 39.68 \\
Qwen3-8B + RL & PPO & 76.80 & 44.97 & 43.80 & 43.03 & 45.62 & 47.43 \\
LongWriter-Zero-32B & SFT + GRPO & 80.66 & 44.69 & 43.66 & 42.40 &  45.21 & 47.48 \\
\rowcolor{lightgreen} Qwen3-4B + R\textsuperscript{2}-Write-SFT + $\text{RL}_\text{p}$ & SFT + PPO & 79.16 & 44.95 & 42.24 & 42.08 & 46.68 & 48.79 \\
\rowcolor{lightgreen} Qwen3-8B + R\textsuperscript{2}-Write-SFT + $\text{RL}_\text{p}$ & SFT + PPO & \textbf{82.00} & \textbf{46.93} & \textbf{44.16} & \textbf{44.03} & \textbf{49.98} & \textbf{49.54} \\
\bottomrule
\end{tabular}
}

\label{tab:main results_appendix}
\end{table*}


\subsection{Can the Model Effectively Use these Patterns?}

To investigate whether the trained model can effectively utilize reflection and revision patterns, we examine all cases where reflection patterns are triggered on WritingBench and DeepResearch Gym benchmarks. Specifically, we compare Qwen3-8B + R\textsuperscript{2}-Write-SFT + $\text{RL}_\text{p}$ against Qwen3-8B + Distill + RL, and categorize triggered cases into three outcomes: Win (our method outperforms baseline), Tie (comparable performance), and Lose (baseline outperforms our method). 

\begin{figure}[ht!]
  \centering
  \vspace{-5pt}
  \resizebox{0.55\textwidth}{!}{
  \includegraphics[width=\textwidth]{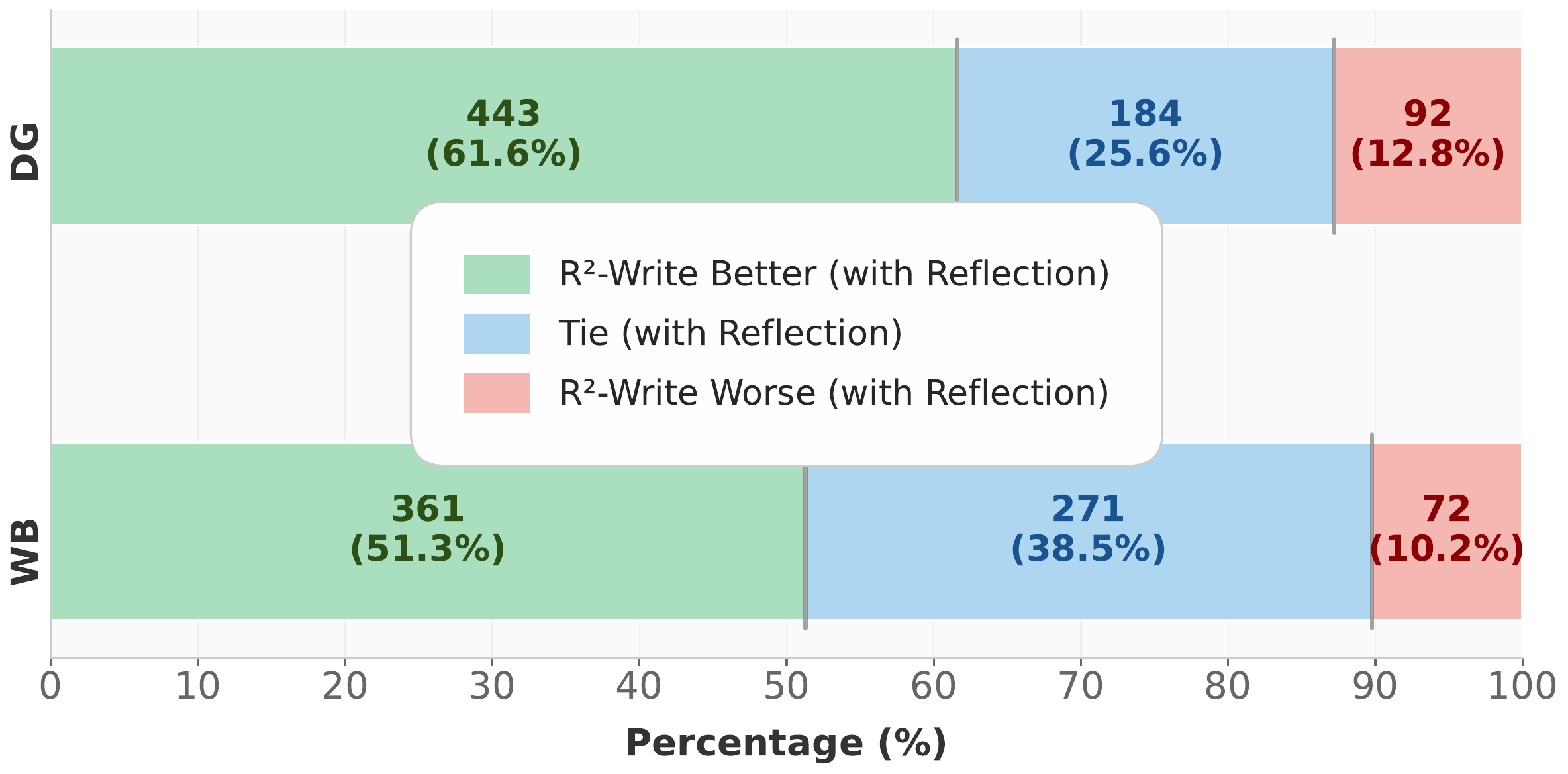}
  }
\caption{
Effectiveness of reflection pattern usage. We categorize cases where reflection patterns are triggered into three outcomes: Win (R\textsuperscript{2}-Write outperforms baseline), Tie (comparable performance), and Lose (baseline outperforms R\textsuperscript{2}-Write). DG represents Deepresearch Gym, and WB means Writing Bench. The vast majority of reflection instances lead to performance improvements, demonstrating that the model effectively leverages these patterns rather than applying them indiscriminately.
}
\label{fig:reflection_analysis}
\end{figure}

Figure~\ref{fig:reflection_analysis} shows the distribution across these categories. We observe that the vast majority of reflection instances (more than 55\%) result in improved performance, while only a small fraction (around 10\%) lead to degradation. This demonstrates that the model has learned to apply reflection and revision patterns effectively and adaptively, rather than indiscriminately, validating the effectiveness of our training approach.


\subsection{Case Study}
\label{sec:case study}
Figures~\ref{fig:reflection_case1} and~\ref{fig:reflection_case2} illustrate how reflection and revision patterns improve generation quality across different task types. For clarity, we manually annotate reflection segments and visualize score improvements; note that these annotations are for illustration purposes and do not reflect the actual model's responses. "Initial average" refers to the preliminary answer before reflection.

In the creative writing task (Figure~\ref{fig:reflection_case1}), the model identifies logical inconsistencies and coherence issues through reflection, then produces a substantially improved response with better narrative structure and argument flow. In the deep research task (Figure~\ref{fig:reflection_case2}), reflection focuses on material utilization and source alignment—the model recognizes insufficient citation coverage and misalignment with provided materials, then systematically addresses these issues through targeted revision that strengthens evidence-based reasoning.

These cases validate that explicit reflection and revision patterns enable models to perform effective iterative self-correction across diverse open-ended writing scenarios.

\subsection{RL Training Dynamics}

We analyze the training dynamics of our RL process to understand how different reward components evolve during training and their impact on model behavior.

\begin{figure}[htbp]
    \centering
    \includegraphics[width=\textwidth]{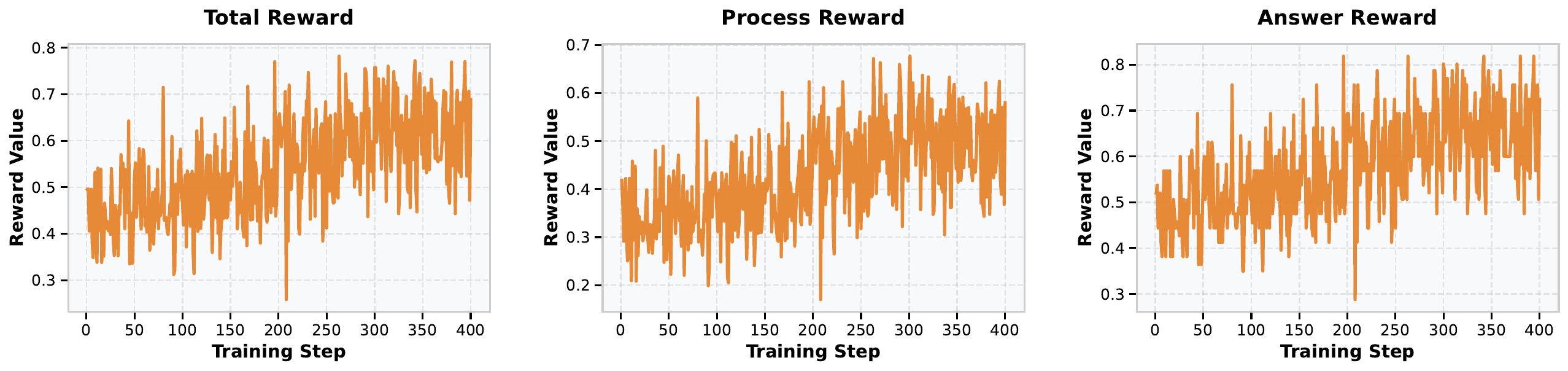}
    \caption{Evolution of total reward, process reward, and answer reward during RL training (steps 0-400). All three reward components show consistent improvement, indicating that the model learns to optimize both the reasoning process and final answer quality simultaneously.}
    \label{fig:rl_rewards}
\end{figure}

\textbf{Reward Evolution.} 
Figure~\ref{fig:rl_rewards} shows the evolution of different reward components during RL training. We observe that the total reward, process reward, and answer reward improve synchronously throughout the training process. This synchronized improvement demonstrates that our model successfully learns to generate better final answers while simultaneously optimizing the thinking process. The consistent upward trend across all three reward metrics indicates that the process reward signal effectively guides the model to develop more coherent and structured reasoning patterns, which in turn leads to higher-quality outputs.

\begin{figure}[htbp]
    \centering
    \includegraphics[width=\textwidth]{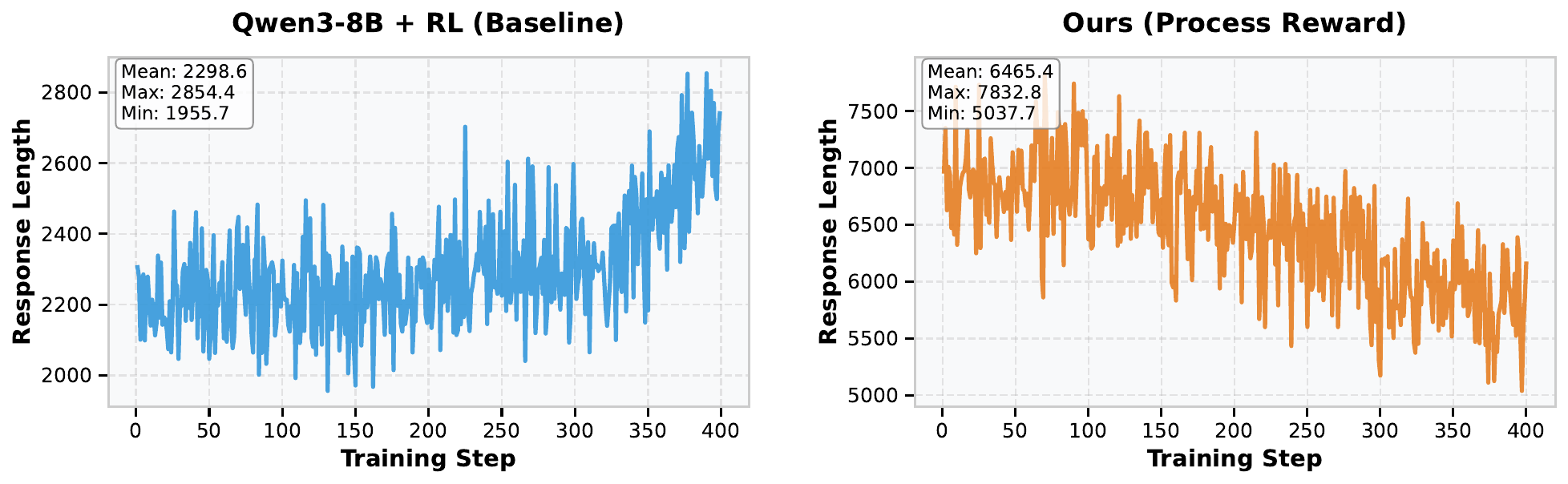}
    \caption{Comparison of response length dynamics during RL training. Our model (right) shows decreasing response length, indicating more efficient reasoning, while the baseline Qwen3-8B + RL (left) exhibits increasing verbosity.}
    \label{fig:response_length}
\end{figure}

\textbf{Response Length Dynamics.} 
Figure~\ref{fig:response_length} presents the response length dynamics for both our model (initialized from R$^2$-Writing SFT) and the baseline Qwen3-8B + RL model. Interestingly, we observe contrasting trends between the two approaches. Our model exhibits a gradual decrease in response length during training, suggesting that the process reward effectively guides the model to generate more concise and efficient reasoning paths while maintaining output quality. This indicates that our approach learns to optimize the reflection and revision process by eliminating redundant steps and focusing on essential reasoning components. In contrast, the baseline Qwen3-8B + RL model shows an increasing trend in response length.

\begin{figure}[htbp]
\centering
\definecolor{darkred}{RGB}{139,0,0}
\definecolor{issuecolor}{RGB}{180,50,50}
\definecolor{beforecolor}{RGB}{100,100,100}
\definecolor{aftercolor}{RGB}{0,100,0}
\begin{promptbox-case}[Case 1: Reflection and Revision in Writing Bench]
\tiny
\begin{lstlisting}[escapeinside={(*}{*)}, basicstyle=\ttfamily\tiny, breaklines=true, linewidth=\textwidth]
(*\textbf{Query:}*) Assume you are the marketing director of a shopping center. Write a complete activity plan for the 8th anniversary celebration. The event lasts one week with a total budget of 500,000 yuan...

(*\textcolor{darkred}{\textbf{Reflection 1: Time Conflict Resolution}}*) (*\textcolor{darkred}{(Event Detail Planning: 5$\rightarrow$8, +3 pts | Execution Feasibility: 5$\rightarrow$7, +2 pts)}*)
(*\textcolor{issuecolor}{\textbf{Issue Identified:}}*)
Wait, I notice a critical time overlap in the June 1st schedule. The celebrity meet-and-greet was set for 14:00-16:00, but the light show spans 12:00-20:00 on the same day. This 2-hour conflict would cause execution chaos as both activities require the same central plaza space and staff resources. The overlapping timeline makes it impossible to properly set up, manage, and transition between events...

(*\textbf{Modification:}*)
(*\textcolor{beforecolor}{\textbf{Before:}}*) "12:00 Anniversary Ceremony (brand representative speech, playing opening documentary of 25 first stores) - 14:00-16:00 Celebrity Meet-and-Greet"
(*\textcolor{aftercolor}{\textbf{After:}}*) "10:00-11:30 Anniversary Ceremony (brand representative speech, playing opening documentary of 25 first stores) - 14:00-16:00 Celebrity Meet-and-Greet - 18:00-20:00 'Flag Gathering Night' Light Show"

(*\textcolor{darkred}{\textbf{Reflection 2: Personnel Configuration Details}}*) (*\textcolor{darkred}{(Event Detail Planning: 8$\rightarrow$9, +1 pt | Execution Feasibility: 7$\rightarrow$8, +1 pt)}*)
(*\textcolor{issuecolor}{\textbf{Issue Identified:}}*)
Looking at personnel allocation again, although the current version mentions daily configuration, it doesn't clarify specific responsibilities for each position. I should supplement with more detailed division of labor. For example, in the celebrity meet-and-greet...

(*\textbf{Modification:}*)
(*\textcolor{beforecolor}{\textbf{Before:}}*) Daily personnel configuration mentioned but lacking specific role responsibilities and detailed breakdown.
(*\textcolor{aftercolor}{\textbf{After:}}*) "Celebrity Meet-and-Greet: 2 hosts (responsible for process control), 10 assistants (assist interactive experience), 20 security guards (maintain order), 10 guides (direct customers)"

(*\textcolor{darkred}{\textbf{Reflection 3: Brand Discount Coverage Gap}}*) (*\textcolor{darkred}{(Event Plan Completeness: 6$\rightarrow$8, +2 pts | Shopping Center Feature Alignment: 7$\rightarrow$9, +2 pts)}*)
(*\textcolor{issuecolor}{\textbf{Issue Identified:}}*)
Now checking the brand discount section. I listed 30 brands, but when carefully cross-checking with the news content, I found that "new stores" mentioned in the news such as Perfect Diary, The Colorist, etc., were not included in the discount list. This may undermine the promotional effect for some new stores and fail to highlight the shopping center's latest offerings, which is a key selling point for the anniversary celebration.

(*\textbf{Modification:}*)
(*\textcolor{beforecolor}{\textbf{Before:}}*) 30 brands listed but missing newly opened stores like Perfect Diary, The Colorist, Meet Fresh mentioned in the background materials.
(*\textcolor{aftercolor}{\textbf{After:}}*) "Add to 'Lifestyle Brands' category: Perfect Diary, The Colorist, Meet Fresh, and other new stores, with discount strength of 30% off popular products"

(*\textcolor{darkred}{\textbf{Reflection 4: AR Technology Implementation Details}}*) (*\textcolor{darkred}{(Execution Feasibility: 8$\rightarrow$10, +2 pts)}*)
(*\textcolor{issuecolor}{\textbf{Issue Identified:}}*)
Regarding digital interactive games, although the current version mentions AR treasure hunt, it lacks technical details and feasibility analysis. Without specific technical implementation methods and backup plans, there's a risk of system failure during the event, which would significantly damage the customer experience and brand reputation. I need to ensure the technology is accessible and has contingency measures.

(*\textbf{Modification:}*)
(*\textcolor{beforecolor}{\textbf{Before:}}*) AR treasure hunt mentioned but no technical specifications, implementation method, or backup plan provided.
(*\textcolor{aftercolor}{\textbf{After:}}*) "Technical Support: Adopt lightweight AR technology, implemented through WeChat mini-program, requiring no special equipment. In case of network issues, can switch to offline QR code scanning card collection backup plan, with 2 dedicated scanning points set up at each store"

(*\textcolor{darkred}{\textbf{Reflection 5: Celebrity Honorarium Budget Adjustment}}*) (*\textcolor{darkred}{(Budget Allocation Rationality: 5$\rightarrow$9, +4 pts)}*)
(*\textcolor{issuecolor}{\textbf{Issue Identified:}}*)
For the celebrity meet-and-greet section, I noticed potential issues with budget allocation. An honorarium of 20,000 yuan per celebrity for 2 celebrities may be too low given current market rates, especially for KOLs with significant following. This is 30-40% below market rates for mid-tier KOLs with 500K-1M followers, risking rejection or forcing us to settle for lower-quality guests that would diminish event appeal and attendance.

(*\textbf{Modification:}*)
(*\textcolor{beforecolor}{\textbf{Before:}}*) "Guest honorarium 40,000 yuan (20,000 yuan per person)"
(*\textcolor{aftercolor}{\textbf{After:}}*) "Guest honorarium 60,000 yuan (30,000 yuan per person), selecting mid-tier KOLs with high interactivity and verified engagement rates above 5% to ensure event participation and audience engagement"

(*\textcolor{darkred}{\textbf{Reflection 6: Digital Interaction Budget Expansion}}*) (*\textcolor{darkred}{(Budget Allocation Rationality: 9$\rightarrow$10, +1 pt)}*)
(*\textcolor{issuecolor}{\textbf{Issue Identified:}}*)
In the budget allocation table, I realized that the 25,000 yuan budget for digital interaction may be insufficient. Currently it only covers mini-program development and specialist fees, but doesn't include technical maintenance (5,000 yuan), server costs (estimated 2,000-3,000 yuan for 3-day high-traffic load), and backup system requirements. This oversight could lead to system crashes during peak usage hours, undermining the entire digital engagement strategy.

(*\textbf{Modification:}*)
(*\textcolor{beforecolor}{\textbf{Before:}}*) "Digital Interaction | 25,000 | 5% | Mini-program development + specialist fees"
(*\textcolor{aftercolor}{\textbf{After:}}*) "Digital Interaction | 35,000 | 7% | Mini-program development (20,000) + specialist fees (10,000) + technical maintenance (5,000) including 24/7 monitoring, server scaling capacity, and emergency backup protocols"

(*\textcolor{darkred}{\textbf{Reflection 7: KPI and Effectiveness Measurement}}*) (*\textcolor{darkred}{(Event Plan Completeness: 8$\rightarrow$10, +2 pts | Event Detail Planning: 9$\rightarrow$10, +1 pt)}*)
(*\textcolor{issuecolor}{\textbf{Issue Identified:}}*)
For the promotional plan section, I noticed the lack of specific KPI settings and effectiveness evaluation methods. Currently it only mentions covering 1 million customer groups, but doesn't clarify how to measure effectiveness. Without specific KPIs, there's no way to: (1) measure campaign effectiveness during execution, (2) justify marketing spend to stakeholders, (3) identify underperforming channels for real-time optimization, or (4) conduct post-event ROI analysis.

(*\textbf{Modification:}*)
(*\textcolor{beforecolor}{\textbf{Before:}}*) "Promotional plan aims to cover 1 million customer groups through multiple channels"
(*\textcolor{aftercolor}{\textbf{After:}}*) "Promotional plan with concrete KPIs: (1) Digital engagement - target interaction rate of 30% (industry benchmark: 15-25%), measured via WeChat mini-program analytics; (2) Conversion metrics - coupon usage rate of 60% (baseline: 40-50%), tracked through unique QR codes; (3) Foot traffic - 20% increase in daily visitors measured via entrance counters with hourly breakdowns"
(*\textbf{Overall Impact Summary:}*)
Through this comprehensive review, I discovered that although the overall framework was solid, it required significant refinement in execution details, budget rationality, and risk control. These modifications make the plan more feasible and effective, ensuring the success of the anniversary celebration.
(*\textbf{Score Progression:}*) Initial Average: 6.6/10 (*$\rightarrow$*) Final Average: 9.4/10 (*\textcolor{darkred}{\textbf{(+2.8 points, +42.4\% improvement)}}*)
\end{lstlisting}
\end{promptbox-case}
\caption{Case 1: Complete Reflection and Revision Process in Writing Bench. }
\label{fig:reflection_case1}
\end{figure}

\begin{figure}[htbp]
\centering
\definecolor{darkred}{RGB}{139,0,0}
\definecolor{issuecolor}{RGB}{180,50,50}
\definecolor{beforecolor}{RGB}{100,100,100}
\definecolor{aftercolor}{RGB}{0,100,0}
\begin{promptbox-case}[Case 2: Reflection and Revision in Deepresearch Gym]
\tiny
\begin{lstlisting}[escapeinside={(*}{*)}, basicstyle=\ttfamily\tiny, breaklines=true, linewidth=\textwidth]
(*\textbf{Query:}*) Is the COVID vaccine dangerous? (materials omitted for brevity)

(*\textcolor{darkred}{\textbf{Reflection 1: Addressing Specific Data Claims}}*) (*\textcolor{darkred}{(Depth: 7$\rightarrow$9, +2 pts)}*)
(*\textcolor{issuecolor}{\textbf{Issue Identified:}}*)
Wait, I notice my report, while well-structured and citing authoritative sources, has a fundamental problem in analyzing vaccine risks. I mainly relied on Johns Hopkins Medical Institution statements emphasizing vaccine safety and benefits, but when reviewing the user-provided research materials, particularly materials 4 and 5, although they contain controversial statements, they do cite specific data points (such as "4000 deaths", "800x more dangerous"). My report completely avoided these data, only vaguely criticizing these sources as "inaccurate" and "not peer-reviewed". This makes my analysis appear one-sided, failing to directly address these specific accusations, weakening the report's persuasiveness and completeness. As a deep research report, I should systematically deconstruct these controversial data rather than simply denying them.


(*\textcolor{aftercolor}{\textbf{After:}}*) "Falsehood: This claim misrepresents data. According to the CDC's Vaccine Safety Data Sheet (VSD) for 2021, the reported death rate for Pfizer and Moderna vaccines in the U.S. was approximately 0.0001-0.0002% (1-2 deaths per million doses), not 400+ per million. The 400,000+ COVID-19 deaths in 2021 (CDC) highlight the stark contrast: vaccination prevents severe outcomes. The smallpox vaccine's historical death rate (0.5-1 per million doses) was not 800x higher than modern vaccines; this comparison is based on flawed data interpretation and ignores contextual differences in vaccine delivery and population health. The CDC explicitly states that VAERS reports (which the Burning Platform references) do not prove causality and require further investigation."

(*\textcolor{darkred}{\textbf{Reflection 2: Long-Term Effects Evidence Precision}}*) (*\textcolor{darkred}{(Clarity: 7$\rightarrow$8, +1 pt)}*)
(*\textcolor{issuecolor}{\textbf{Issue Identified:}}*)
Wait, I notice that when discussing long-term side effects, I cited material 3 (UAB Health System) stating "no long-term chronic problems found", but material 3 itself is a brief press release without specific data or research details. I should be more cautious in my phrasing to avoid over-generalization. The current version lacks explanation of the "long-term impact" research status, which may mislead readers into thinking the risks are completely unknown. A more nuanced presentation would acknowledge ongoing monitoring while accurately representing current findings.

(*\textbf{Modification:}*)
(*\textcolor{beforecolor}{\textbf{Before:}}*) "No Evidence of Long-Term Harm: Over 6 months of real-world data shows no increase in chronic conditions (e.g., autoimmune diseases, fertility issues) linked to vaccination (UAB, 2021)."
(*\textcolor{aftercolor}{\textbf{After:}}*) "No Evidence of Long-Term Harm: Over 6 months of real-world data and ongoing studies (e.g., CDC's VSD and VAERS) have not identified significant chronic conditions (e.g., autoimmune diseases, fertility issues) linked to vaccination. The UAB Health System (2021) notes that short-term side effects are well-documented, but long-term data continues to be collected as vaccines have been in widespread use for over two years. No credible evidence links vaccines to chronic health issues beyond the well-managed rare side effects."

(*\textcolor{darkred}{\textbf{Reflection 3: Spike Protein Mechanism Clarification}}*) (*\textcolor{darkred}{(Depth: 9$\rightarrow$9 maintained | Breadth: 9$\rightarrow$9 maintained | Insightfulness: 7$\rightarrow$8, +1 pt)}*)
(*\textcolor{issuecolor}{\textbf{Issue Identified:}}*)
Looking back at the "Debunking Misinformation" section, although I refuted the claims that "vaccines cause infertility" and "spike protein is toxic", material 5 cited Dr. Brindle's statement claiming spike protein accumulates in blood and causes tissue damage. I should respond to this point in more detail, citing scientific literature or institutional positions, rather than relying solely on Johns Hopkins. The current version lacks in-depth explanation of mRNA vaccine mechanisms, which may not thoroughly eliminate readers' doubts. More comprehensive scientific consensus is needed.

(*\textbf{Modification:}*)
(*\textcolor{beforecolor}{\textbf{Before:}}*) Basic debunking of spike protein toxicity claims without detailed mechanistic explanation or additional authoritative sources.
(*\textcolor{aftercolor}{\textbf{After:}}*) Added: "Additional scientific consensus: The American College of Obstetricians and Gynecologists (ACOG) and the Society for Maternal-Fetal Medicine confirm that COVID-19 vaccines do not affect fertility in women, including those trying to conceive (2021). Regarding the spike protein, it is a harmless fragment produced temporarily in muscle cells by mRNA vaccines to train the immune system. It does not enter the nucleus, alter DNA, or persist in the body long-term (Johns Hopkins, 2022). Studies showing spike protein distribution (e.g., Pfizer's biodistribution data) indicate it is present in various tissues but at concentrations far below harmful levels, and it is rapidly degraded (as confirmed by independent virology research)."

(*\textcolor{darkred}{\textbf{Reflection 4: Fact vs. Interpretation Distinction}}*) (*\textcolor{darkred}{(Clarity: 8$\rightarrow$9, +1 pt | Balance: 9 maintained)}*)
(*\textcolor{issuecolor}{\textbf{Issue Identified:}}*)
On second thought, my report, though structurally clear, does not clearly distinguish between "factual data" and "controversial interpretation", which may cause readers to confuse scientific consensus with conspiracy theories. I should add a section that directly compares specific accusations in materials 4/5 with authoritative institutional data. For example, in the "blood clot risk" section, clearly explain the gap between J&J vaccine's TTS incidence rate (1-2 per million doses) and material 4's claim of "800x more dangerous", citing CDC's latest guidelines.

(*\textbf{Modification:}*)
(*\textcolor{beforecolor}{\textbf{Before:}}*) General discussion of blood clot risks without direct point-by-point comparison to specific misinformation claims.
(*\textcolor{aftercolor}{\textbf{After:}}*) Added subsection in "Real-World Safety Data": "Addressing Specific Misinformation: Blood Clots and Myocarditis - The Burning Platform claims J&J vaccine causes '800x more deaths,' but CDC data shows TTS cases are rare (1-2.7 per million doses in 2021), with no deaths attributed to the condition. Moderna and Pfizer vaccines have negligible TTS risk. - Material 5's claim of '52,000 deaths in Europe' is not supported by EU pharmacovigilance reports (2021), which show no excess mortality linked to vaccines. Citation: [CDC. (2021). Update to the Emergency Use Authorization for Janssen COVID-19 Vaccine.]"

(*\textcolor{darkred}{\textbf{Reflection 5: Booster and Long-Term Monitoring Context}}*) (*\textcolor{darkred}{(Insightfulness: 8$\rightarrow$9, +1 pt)}*)
(*\textcolor{issuecolor}{\textbf{Issue Identified:}}*)
Finally, I realize the report does not adequately discuss vaccine boosters and the necessity of long-term monitoring, which may affect the comprehensive assessment of "risk balance". The current version lacks explanation of variant impact and immune durability, which should be supplemented to enhance depth and provide more forward-looking insights for readers concerned about evolving pandemic dynamics.

(*\textbf{Modification:}*)
(*\textcolor{beforecolor}{\textbf{Before:}}*) Conclusion section lacks discussion of ongoing monitoring, variant evolution, and booster recommendations.
(*\textcolor{aftercolor}{\textbf{After:}}*) Added to "Conclusion" section: "Ongoing monitoring remains critical as new variants emerge (e.g., Omicron) and vaccine efficacy may wane over time, leading to booster recommendations. The CDC and FDA continuously update guidelines based on real-world data, ensuring safety profiles remain robust against evolving threats."

(*\textbf{Overall Impact Summary:}*)
Through these systematic reflections, I transformed the report from a well-structured but potentially one-sided safety endorsement into a comprehensive, evidence-based analysis that directly engages with controversial claims while maintaining scientific rigor. The revisions enhance the report's credibility by demonstrating thorough engagement with dissenting voices and providing granular data comparisons.
(*\textbf{Score Progression:}*) Initial Average: 7.7/10 (*$\rightarrow$*) Final Average: 8.8/10 (*\textcolor{darkred}{\textbf{(+1.1 points, +14.3\% improvement)}}*)
\end{lstlisting}
\end{promptbox-case}
\caption{Case 2: Complete Reflection and Revision Process in Writing Bench}
\label{fig:reflection_case2}
\end{figure}

\begin{figure}[htbp]
\centering
\begin{promptbox}[Prompt for Generating Query-Specific Evaluation Rubrics]
\begin{lstlisting}
You are an expert evaluator tasked with creating precise scoring rubrics. 

Given the following:
Query: {query}

## Task Workflow

### Step 1: Internal Analysis (Not included in output)
Analyze the query to identify:
- **Explicit Requirements**: Directly stated demands in the query
- **Implicit Requirements**: Inferred needs based on context and domain conventions (e.g., originality, format compliance, target audience appropriateness and so on.)

**Key Point Extraction Principles:**
- **Comprehensiveness**: Capture all explicit and implicit requirements without omission
- **Appropriate Granularity**: Consolidate related points to avoid redundancy while preserving clarity (aim for 5-10 key points for typical queries)
- **Generalizability**: Focus on requirements applicable across valid responses.

### Step 2: Define Scoring Criteria
Establish detailed scoring rules for each key point using the following framework:

**Scoring Standards:**
- **Fully Correct** (+1 point): The key point is mentioned and expressed accurately
- **Partially Correct** (+0.5 point): The key point is mentioned but the answer is incomplete or partially correct (adjust based on completeness)
- **Incorrect or Not Mentioned** (0 points): The key point is mentioned but contains factual errors (e.g., data errors, causal inversions, logical errors), or is not mentioned at all

### Step 3: Output Format Requirements
Generate evaluation rubrics in the following JSON format:
- **key_point**: The detailed description of the key point
- **score_standards**: The scoring standards for each key point
- **Language consistency**: Use the same language as the query

{
  "rubrics": [
    {
      "id": 1, 
      "key_point": "Description of the first key evaluation point",
      "score_standards": "Fully correct: +1 point. Partially correct: +0.5 point. Not mentioned or incorrect: 0 points."
    },
    {
      "id": 2,
      "key_point": "Description of the second key evaluation point", 
      "score_standards": "Fully correct: +1 point. Partially correct: +0.5 point. Not mentioned or incorrect: 0 points."
    }
  ]
}

Important: Only output the JSON object, nothing else.
\end{lstlisting}
\end{promptbox}
\caption{Prompt for generating query-specific evaluation rubrics for language model responses.}
\label{fig:rubric_generation}
\end{figure}

\begin{figure}[htbp]
\centering
\begin{promptbox}[Prompt for Generating General-Quality Evaluation Criteria]
\begin{lstlisting}
You are an evaluation assistant. Given a query, generate appropriate evaluation criteria in JSON format.

Query: {query}

## Task Requirements

1. **Analyze Query Type**: Identify the task category (e.g., creative writing, technical explanation, business strategy, instructional content, report generation, etc.)

2. **Select Relevant Criteria**: Choose 6-8 criteria from these dimensions based on query type (including but not limited to):
   - Clarity: Organization and communication effectiveness
   - Completeness: Coverage of required aspects
   - Depth: Level of detail and thoroughness
   - Creativity: Originality and innovation
   - Appropriateness: Fit to task, audience, and constraints
   - Practical Value: Actionability and usefulness
   - Insight: Non-obvious perspectives and understanding
   - Factual Accuracy: Correctness of facts and data
   - Technical Correctness: Precision in technical/scientific content
   - Visual Structure: Effective use of tables, lists, diagrams
   - Evidence Alignment: Consistency between claims and provided materials

3. **Adaptation Guidelines by Task Type** (including but not limited to):
   - **Creative tasks**: Emphasize Creativity, Appropriateness; de-emphasize Factual Accuracy
   - **Technical/Scientific tasks**: Emphasize Accuracy, Technical Correctness, Depth
   - **Instructional tasks**: ...
   - **Business/Strategy tasks**: ...
   - **Report generation**: ...

4. **Criterion Description Format**: For each criterion, specify what constitutes a strong answer using 2-4 concrete evaluation points in the format: "A strong answer should: 1)..., 2)..., 3)..."

## Output Format Example

{
  "EVAL_CRITERIA": [
    {
      "name": "Clarity",
      "description": "Evaluate how clearly the answer communicates its ideas and organization. A strong answer should: 1) have a logical structure appropriate to the task; 2) avoid substantial redundancy or overlapping concepts; 3) use precise language suited to the audience."
    },
    {
      "name": "Completeness",
      "description": "Evaluate whether the answer addresses all key aspects required. A strong answer should: 1) cover main components implied by the task; 2) not omit important elements; 3) leave no sense of major pieces missing."
    },
    ...
  ]
}

Important: Only output the JSON object, nothing else.
\end{lstlisting}
\end{promptbox}
\caption{Prompt for generating general-quality evaluation criteria.}
\label{fig:adaptive_criteria}
\end{figure}

\begin{figure}[htbp]
\centering
\begin{promptbox}[Prompt for constructing Self-Reflection and Revision during Data Construction]
\begin{lstlisting}
You are a professional writing assistant. Your task is to analyze the article and identify areas for improvement based on the comprehensive evaluation feedback, and then reconstruct a detailed self-verification and error-discovery thought process.

**Important: Output with the same language as the [Original Article]** If the [Topic] and [Original Article] use Chinese, you must output in Chinese.

[Topic]
{question}
[Original Article]
{original_article}
[Evaluation Feedback]

The evaluation consists of two complementary parts:

<evaluation_part_1: Overall Article Quality>
This section evaluates the article holistically across multiple dimensions including writing depth, clarity of expression, logical coherence, structural organization, and overall fluency.
{feedback_quality}
</evaluation_part_1>

<evaluation_part_2: Query-Specific Requirements>
This section assesses how well the article addresses the specific requirements and key points outlined in the original topic/query.
{feedback_kpr}
</evaluation_part_2>

## Task Instructions
Carefully analyze the original article using the information above, and reconstruct a natural, internal monologue of self-checking, as if you are verifying your own work, discovering problems, and deciding how to revise.
Your output must read like an authentic self-reflection, as if YOU are the original author reviewing your own article.
**Key Requirements:**
1. **Reflect and Discover Problems**
   - Use the information from the evaluation to identify problems, but never explicitly mention external feedback. Frame all discoveries as your own insights.
   - Always write as if YOU authored the article and are now reviewing it.
   - Use reflective phrases naturally:
     * English: "Wait, I notice that...", "Looking back at...", "On second thought...", "This section seems to...", "I should have..."
2. **Describe Revision Ideas**
   - For each problem, explain:
     * What the problem is
     * Why it is problematic
     * How you plan to fix it
3. **Provide Concrete Revision Content**
   - For every issue, provide specific before-and-after text.
   - Use format: "I will revise the original '[exact original text]' to '[specific new version]'"
   - Provide complete, usable revision content.
4. **Output Style**
   - Write as a continuous internal monologue, not rigid sections.
   - Do NOT use headings like "Problem 1", "Solution".
   - Output entirely in the same language as the [Original Article] and [Topic].
Begin your critical self-analysis now:
\end{lstlisting}
\end{promptbox}
\caption{Prompt for constructing self-reflection and revision thought process during Data Construction.}
\label{fig:self_revision_1}
\end{figure}

\begin{figure}[htbp]
\centering
\begin{promptbox}[Prompt for Response Revision Based on Self-Reflection during Data Construction]
\begin{lstlisting}
You are a professional writing assistant. Your task is to revise the article based on the critical self-reflection provided.

[Topic]
{question}

[Original Article]
{original_article}

[Evaluation Feedback]

The evaluation consists of two complementary parts:

<evaluation_part_1: Overall Article Quality>
This section evaluates the article holistically across multiple dimensions including writing depth, clarity of expression, logical coherence, structural organization, and overall fluency.

{feedback_quality}
</evaluation_part_1>

<evaluation_part_2: Query-Specific Requirements>
This section assesses how well the article addresses the specific requirements and key points outlined in the original topic/query.

{feedback_kpr}
</evaluation_part_2>

[Critical Self-Reflection]
{self_reflection}

## Revision Instructions

Based on the critical self-reflection above, please produce the revised article that addresses all the identified issues and planned improvements.

**Requirements:**
- Include ONLY the final revised article content
- Do NOT include any explanations, labels, metadata, or commentary
- Ensure all improvements identified in the self-reflection are naturally integrated
- Maintain coherent flow and readability throughout

**Output Format:**
- Direct output of the revised article content only.
- **Please note that the language must be determined according to the language of the [Original Article]**

Begin your revision now:
\end{lstlisting}
\end{promptbox}
\caption{Prompt for response revision based on self-reflection during Data Construction.}
\label{fig:article_revision}
\end{figure}

\begin{figure*}[ht]
\centering
\begin{promptbox}[Generation Prompt for Writing Tasks]
\textbf{System:} Please reason step by step, and put your final answer within \texttt{\$\textbackslash boxed\{\}\$}.

\textbf{User:} \{Question\}

\textbf{Assistant:} 
\end{promptbox}
\caption{The generation prompt for Math tasks in our experiments.}
\label{fig:prompt_eval_math}
\end{figure*}

\begin{figure*}[ht]
\centering
\begin{promptbox}[Generation Prompt for Math Tasks]
\textbf{System:} You are a helpful assistant. Please provide a detailed response to the following writing task.

\textbf{User:} \{Question\}

\textbf{Assistant:} 
\end{promptbox}
\caption{The Generation Prompt on Writing tasks (WritingBench and HelloBench) in our experiments.}
\label{fig:prompt_eval_writing}
\end{figure*}

\begin{figure}[htbp]
\centering
\begin{promptbox}[Prompt for Answer Reward in RL training]
\begin{lstlisting}
Please act as an impartial judge and evaluate the quality of the responses provided by two AI assistants to the user question displayed below. You should choose the assistant that follows the user's instructions and answers the user's question better. Your evaluation should consider the following dimensions.

{criteria}

Begin your evaluation by comparing the two responses and provide a short explanation. Avoid any position biases and ensure that the order in which the responses were presented does not influence your decision. Do not allow the length of the responses to influence your evaluation. Do not favor certain names of the assistants. Be as objective as possible. After providing your explanation, output your final verdict by strictly following this format: "[[A]]" if assistant A is better, "[[B]]" if assistant B is better, and "[[C]]" for a tie. NOTE: If the response contains severe repetition or redundancy, it should be viewed as low quality score, losing the comparison.

[User Question]
{question}

[The Start of Assistant A's Answer]
{answer_a}
[The End of Assistant A's Answer]

[The Start of Assistant B's Answer]
{answer_b}
[The End of Assistant B's Answer]
\end{lstlisting}
\end{promptbox}
\caption{Prompt for Answer Reward in RL training}
\label{fig:answer_reward}
\end{figure}

\begin{figure}[htbp]
\centering
\begin{promptbox}[Prompt for Process Reward in RL training (Step 1)]
\begin{lstlisting}
You are a professional thinking process analyst, skilled at identifying and extracting verification steps in reasoning processes.
## Task Description
Below is a writing task (query) and the user's complete response, which contains two parts:
- **thinking_process**: The thought process
- **final_answer**: The final answer
## Your Task
Please carefully analyze the `thinking_process`, identify and extract **all verification processes** within it.

**Definition of verification process**:
- Checking against the original query requirements
- Reviewing previous thoughts or answers
- Discovering errors, omissions, or deficiencies
## Input Content

<query>
{query}
</query
<thinking_process>
{thinking_process}
</thinking_process>

<final_answer>
{final_answer}
</final_answer>
## Output Requirements
Please output all identified verification processes in **JSON format**, each verification point should include:
| Field | Description |
|-------|-------------|
| `id` | Serial number of the verification point (starting from 1) |
| `content` | Detailed text fragment of the verification process (quoted from original), including reflection and problem discovery |

## Output Format Example

```json
{
  "verifications": [
    {
      "id": 1,
      "content": "Wait, looking back at the previous analysis, I notice that..."
    },
    {
      "id": 2,
      "content": "Hold on, there's a logical issue here..."
    }
  ],
  "total_count": 2
}
Important: Only output the JSON object, nothing else.
\end{lstlisting}
\end{promptbox}
\caption{Prompt for Process Reward in RL training (Step 1).}
\label{fig:process_reward_step1}
\end{figure}

\begin{figure}[htbp]
\centering
\begin{promptbox}[Prompt for Process Reward in RL training (Step 2)]
\tiny{
\begin{lstlisting}
## Task Background
Below is a writing task (query) and the user's complete response, which contains two parts:
- **thinking_process**: The thought process
- **final_answer**: The final answer
## Core Task
Please evaluate the quality of **all verification processes** in the `thinking_process`.
### Definition of Verification Process
A verification process refers to self-checking steps that appear in the thinking process, specifically manifested as: Checking against the original query requirements Reviewing previous thoughts or answer content Discovering errors, omissions, or deficiencies
---
## Evaluation Dimensions
Please conduct **three-dimensional scoring** for each verification process:
### Dimension 1: Validity of Issue Discovery
> Based on the rubrics, judge whether the identified issue is valuable
| Score | Standard | Description |
|:-----:|----------|-------------|
| **+1** | Issue is reasonable and valuable | The discovered problem indeed exists and substantially helps improve answer quality. Should be judged in conjunction with the rubrics. |
| **-1** | Issue is trivial or unreasonable | The problem does not exist, or is illogical, or is over-interpreted |
### Dimension 2: Quality of Correction Suggestion
> Warning: **Prerequisite**: Only evaluate this dimension when Dimension 1 score > 0, otherwise this dimension scores 0
| Score | Standard | Description |
|:-----:|----------|-------------|
| **+1** | Correction aligns with rubrics | Improvement direction is basically consistent with the rubrics |
| **-1** | Correction is inappropriate | Violates rubrics requirements, or lacks specific modification description |
### Dimension 3: Implementation in Answer
> Check whether `final_answer` actually adopts the correction suggestions from `thinking_process`
| Score | Standard | Description |
|:-----:|----------|-------------|
| **+1** | Correction implemented | final_answer reflects the improvements proposed in thinking |
| **-1** | Correction not implemented | final_answer does not reflect the corresponding improvements |
## Input Content
<query>
{query}
</query>
<final_answer>
{final_answer}
</final_answer>
<rubrics>
{rubrics}
</rubrics>
Please refer to the query and thinking_process above, and evaluate the following verifications from the three dimensions mentioned above.
<verifications>
{verifications}
</verifications>
## Output Requirements
Please output the evaluation results in **JSON format**
\end{lstlisting}}
\end{promptbox}
\caption{Prompt for Process Reward in RL training (Step 2).}
\label{fig:process_reward_step2}
\end{figure}

\begin{figure}[htbp]
\centering
\begin{promptbox}[Prompt for Analyzing Patterns of Thinking Trajectories]
\begin{lstlisting}
You are a helpful assistant. The following is a chain-of-thought produced by a language model in response to a math & science problem:
Question: {question}
Reasoning: {reasoning}
Ground Truth: {ground_truth} (Evaluation Rubrics {Evaluation Rubrics})
Analyze the reasoning for the following patterns:
Pattern 1: Answer Verification
Determine whether the reasoning includes any explicit or implicit answer verification steps - moments where the model checks intermediate computations or final results for correctness.
Example: "Let's verify this result by...", "Checking: 5 + 3 = 8 checkmark"
- Identify each distinct answer verification step and extract the specific content.
Pattern 2: Backtracking Behavior
Determine whether the reasoning demonstrates backtracking - where the model identifies an error or dead end and switches to a different approach.
Example: "This approach won't work because..., let's try another method...", "Wait, I made an error. Let me recalculate..."
- Identify each distinct backtracking instance and extract the specific content.
Pattern 3: Subgoal Setting
Determine whether the reasoning includes any explicit subgoals - intermediate steps that break the problem into smaller, manageable parts.
Example: "First, I'll calculate..., then I'll...", "Step 1: Find the area. Step 2: Calculate the perimeter."
- Identify each clearly defined subgoal and extract the specific content.
Pattern 4: Backward Chaining
Determine whether the reasoning includes backward chaining - starting from the target result and reasoning backward to infer inputs or steps.
Example: "To get 24, I need 24 / 2 = 12...", "Working backwards from the answer..."
- Identify each distinct backward chaining attempt and extract the specific content.
Pattern 5: Summarization
Determine whether the reasoning includes summarization - identifying completed subtasks, summarizing progress, and determining the next steps.
Example: "Now we have obtained x = 5, next we need to...", "So far we've found..., the remaining step is..."
- Identify each summarization instance and extract the specific content.
For each task:
1. List all instances found with their specific content from the reasoning
2. For each instance, evaluate contribution:
   - "yes": The final answer matches the ground truth AND this specific instance meaningfully contributed to reaching the correct answer
   - "no": The final answer matches the ground truth BUT this specific instance did not meaningfully contribute to the correct answer
   - "na": The final answer does not match the ground truth
3. Provide the total count of instances found
Please output your analysis in the following JSON format:
{
    "answer_verification": {"instances": [{"id": 1, "content": "...", "contribution": "yes/no/na"}], "count": <int>},
    "backtracking": {"instances": [...], "count": <int>},
    "subgoal_setting": {"instances": [...], "count": <int>},
    "backward_chaining": {"instances": [...], "count": <int>},
    "summarization": {"instances": [...], "count": <int>}
}
Important: Only output the JSON object, nothing else.
\end{lstlisting}
\end{promptbox}
\caption{Prompt for analyzing patterns of thinking trajectories in language model outputs. For writing tasks, we replace the \texttt{Ground Truth} in the prompt with \texttt{Evaluation Rubrics}.}
\label{fig:pattern_classification}
\end{figure}

\begin{figure}[htbp]
\centering
\begin{promptbox}[Prompt for Classifying Revision Types in Writing Tasks]
\begin{lstlisting}
Analyze the reasoning content to identify reflection and revision patterns, then classify those that contributed to improving the writing quality.
**Input:**
Question/Task: {question}
Reasoning Content: {reasoning_content}
Evaluation Rubrics: {rubrics}

**Step 1: Extract Reflection and Revision Patterns**
Identify all instances where the model:
- Recognizes issues or gaps in the current output
- Proposes corrections or improvements
- Revises the approach or content

**Step 2: Evaluate Contribution**
For each identified pattern, determine if it meaningfully contributed to improving the writing quality according to the evaluation rubrics. Only include patterns that had a positive impact.

**Step 3: Classify Contributing Patterns**
Classify each contributing pattern into ONE of the following categories:
1. **Requirement Alignment (RA)**: Corrections that align output with user's explicit requirements
   - Addressing missing key elements/sections that user requested
   - Fixing format violations (word count, structure, style)
   - Adjusting scope or target audience to match specifications
   - Note: Any revision that brings the output closer to user's explicit instructions.
   
2. **Factual & Logical Correction (FLC)**: Corrections of errors in facts, data, or reasoning
   - Fixing incorrect numbers, calculations, or citations
   - Correcting legal articles, historical events, or technical principles
   - Resolving logical contradictions or flawed reasoning
   - Updating outdated information or misquoted sources
   - Note: Any revision that corrects factual inaccuracies or logical flaws.
   
3. **Quality Enhancement (QE)**: Improvements to overall writing quality
   - Adding missing details, examples, or deeper analysis
   - Improving language clarity and terminology precision
   - Strengthening theoretical support and depth
   - Enhancing coherence, flow, and readability
   - Adjusting tone, style, or formatting for professionalism
   - Note: Any improvement to writing quality without fixing factual errors or requirement mismatches.

**Response Format:**
{
    "patterns": [
        {
            "id": 1,
            "content": "Brief description of the identified pattern",
            "category": "RA/FLC/QE",
            "contribution": "How this pattern improved the writing quality"
        }
    ],
}
**Important:** 
- Only include patterns that positively contributed to quality improvement
- Output ONLY the JSON object, nothing else
\end{lstlisting}
\end{promptbox}
\caption{Prompt for classifying revision types in writing task reflection. The model analyzes the reasoning content to identify the primary issue: Requirement Alignment (RA), Factual \& Logical Correction (FLC), or Quality Enhancement (QE).}
\label{fig:revision_classification}
\end{figure}

\end{document}